\RequirePackage{fix-cm}
\documentclass[journal]{IEEEtran}
 
\usepackage{graphicx}
\usepackage{times}
\usepackage{epsfig}
\usepackage{graphicx}
\usepackage{amsmath}
\usepackage{amssymb}
\usepackage{bm}
\usepackage{multirow}

 \usepackage{caption}
 \usepackage{subcaption}
 \usepackage{bbding}
\begin{document}

\title{A comprehensive survey on point cloud registration}

\author{Xiaoshui Huang$^{[1]}$, Guofeng Mei$^{[2]}$, Jian Zhang$^{[2]}$, Rana Abbas$^{[1]}$ \\
	{[1] The University of Sydney, [2] University of Technology Sydney}\\

	{\tt\small Xiaoshui.Huang@sydney.edu.au, {Guofeng.Mei@student.uts.edu.au, Jian.Zhang@uts.edu.au}, rana.abbas@sydney.edu.au}
}


\date{Received: date / Accepted: date}

\maketitle

\begin{abstract}
Registration is a transformation estimation problem between two point clouds, which has a unique and critical role in numerous computer vision applications. The developments of optimization-based methods and deep learning methods have improved registration robustness and efficiency. Recently, the combinations of optimization-based and deep learning methods have further improved performance. However, the connections between optimization-based and deep learning methods are still unclear. Moreover, with the recent development of 3D sensors and 3D reconstruction techniques, a new research direction emerges to align cross-source point clouds. This survey conducts a comprehensive survey, including both same-source and cross-source registration methods, and summarize the connections between optimization-based and deep learning methods, to provide further research insight. This survey also builds a new benchmark to evaluate the state-of-the-art registration algorithms in solving cross-source challenges. Besides, this survey summarizes the benchmark data sets and discusses point cloud registration applications across various domains. Finally, this survey proposes potential research directions in this rapidly growing field. 
\end{abstract}

\section{Introduction}
Point cloud has become the primary data format to represent the 3D world as the fast development of high precision sensors such as LiDAR and Kinect. Because the sensors can only capture scans within their limited view range, the registration algorithm is required to generate a large 3D scene. Point cloud registration is a problem to estimate the transformation matrix between two-point cloud scans. Applying the transformation matrices, we can merge the partial scans about the same 3D scene or object into a complete 3D point cloud. 

The value of point cloud registration is its unique and critical role in numerous computer vision applications. Firstly, \textbf{3D reconstruction}. Generating a complete 3D scene is a basic and significant technique for various computer vision applications, including high-precision 3D map reconstruction in autonomous driving, 3D environment reconstruction in robotics and 3D reconstruction for real-time monitoring underground mining.  For example, registration could construct the 3D environment for route plan and decision-making in robotics applications. Another example could be a large 3D scene reconstruction in the underground mining space to monitor mining safety accurately. Secondly, \textbf{3D localization}. Locating the position of the agent in the 3D environment is particularly important for robotics. For example, a driverless car estimates its position on the map (e.g. $<10cm$) and its distance to the road's boundary line. Point cloud registration could accurately match a current real-time 3D view to its belonging 3D environment to provide a high-precision localization service. This application shows that the registration provides a solution to interact with the 3D environment for an autonomous agent (e.g. robots or drive-less car). Thirdly, \textbf{pose estimation}. Aligning a point cloud $A$ (3D real-time view) to another point cloud $B$ (the 3D environment) could generate the pose information of point cloud $A$ related to point cloud $B$. The pose information could be used for decision-making in robotics. For example, the registration could get the robotics arm's pose information to decide where to move to grab an object accurately. The pose estimation application shows that the registration also provides a solution to know the environment's agent information.  Since point cloud registration plays a critical role in numerous valuable computer vision applications, there is a significant urgent need to conduct a comprehensive survey of the point cloud registration to benefit these applications.

The registration problem has endured thorough investigation from optimization aspects \cite{Bernard_2018_CVPR,bes1992method,dym2017ds++,forstner2017efficient,iglesias2020global,kezurer2015tight,le2019sdrsac,segal2009generalized,yang2019polynomial}.  Most of the existing registration methods are formulated by minimizing a geometric projection error through two processes: correspondence searching and transformation estimation. These two processes alternatively conduct until the geometric projection error is minimum. Upon the accurate correspondences known, the transformation estimation has a close-form solution \cite{bes1992method}. 

Recently, there are many development in 3D deep learning techniques \cite{3dmatch,deng20193d,choy2019fully,yang2020learning,valsesia2020learning}. These techniques aim to extract distinctive features for 3D points and find accurate correspondences. Then, these correspondences are used to estimate a transformation with a separate transformation estimation stage. 
There is also some combination of conventional registration optimization strategies and deep learning techniques in an end-to-end framework \cite{huang2020feature,choy2020deep,aoki2019pointnetlk,wang2019deep}. Their experiments show a significant performance gain. However, the connections between optimization-based and deep learning methods are still unclear.

Moreover, there is an emerging topic about cross-source point cloud registration with the development of 3D sensors, such as Kinect and Lidar. Each 3D sensor has its distinct advantages and limitations. For example, Kinect can generate dense point clouds, while the view range is usually limited to 5 meters. Lidar has a long view range while generating sparse point clouds. Data fusion of these different kinds of 3D sensors combines their advantages and is a cross-source point cloud registration problem \cite{huang2017coarse,huang2017systematic,huang2016coarse}. The cross-source point cloud registration has wide applications such as building construction, augmented reality, and driverless vehicles. For example, the builders compare the 3D CAD model with real-time LiDAR scans to evaluate the contract's current construction quality.  The development in both same-source and cross-source point cloud registration also requires a comprehensive survey to summarize the recent advances.

Although there are a few existing reviews on point cloud registration \cite{cheng2018registration,pomerleau2015review,saiti2020application}, they mainly focus on the view of conventional point cloud registration. \cite{zhang2020deep} surveys deep learning techniques. However, the recent development of cross-source point cloud registration has not been surveyed, and the connections between conventional optimization and recent deep learning methods are unclear. To stimulate point cloud registration development in industrial and academic, we conduct a comprehensive survey by summarizing the recent fast development of point cloud registration (1992-2021), including both same-source and cross-source, conventional optimization and current deep learning methods. Moreover, we summarize the connections between optimization strategies and deep learning techniques.  

Besides, while the recent deep learning-based registration techniques achieve high accuracy on same-source point cloud databases, cross-source point clouds' performance is less reported. This survey will build a benchmark to evaluate the recent state-of-the-art registration algorithms on a cross-source dataset.

\begin{figure*}
	\centering
	\includegraphics[width=0.8\linewidth]{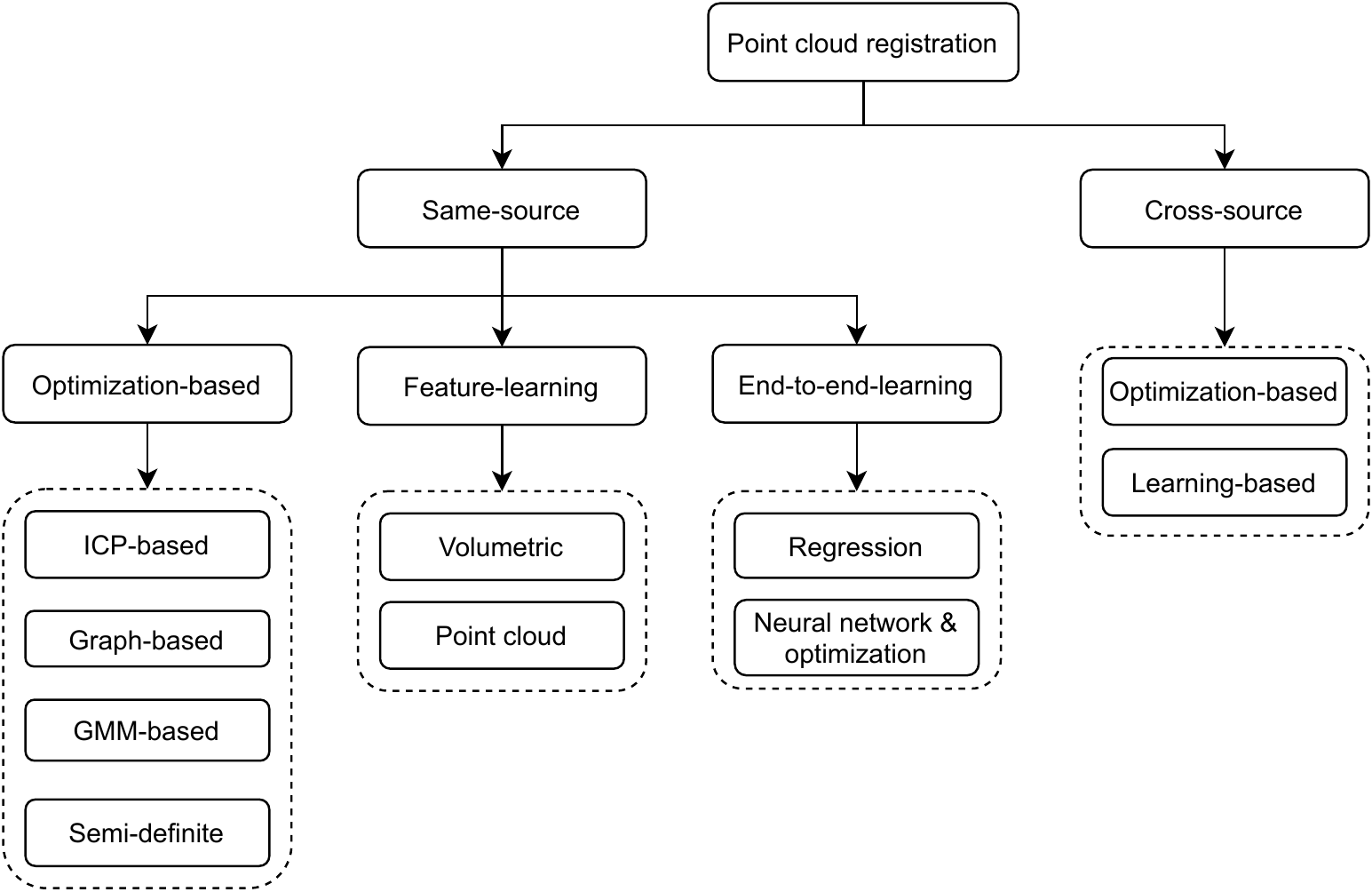}
	\caption{Taxonomy of point cloud registration}
	\label{f_category}
\end{figure*}

\textbf{Our contributions.} Our paper makes notable contributions summarized as follows:

\begin{itemize}
	\item \textbf{Comprehensive review}. We provide the most comprehensive overview for same-source point cloud registration, including conventional optimization and modern deep learning methods (1992-2021). We summarize the challenges, analyze the advantages and limitations of each category of registration methods. Moreover, the connections between conventional optimization and modern deep learning methods are summarized in this paper. These connections could provide insights for future research.
	\item \textbf{Review of cross-source registration}. For the first time, we provide a literature review about cross-source point cloud registration. This survey provides insights for data fusion research from different 3D sensors (e.g., Kinect and Lidar). Figure \ref{f_category} shows a taxonomy of point cloud registration.
	\item \textbf{New comparison}. We build a novel cross-source point cloud benchmark. Then, the existing state-of-the-art registration algorithms' performance is evaluated and compared on the new cross-source point cloud benchmark. This survey can provide a guide for choosing and developing new registration approaches for cross-source point cloud applications.
	\item \textbf{Applications and future directions.} We summarize the potential applications of point cloud registration and explore the research directions in real applications. Besides, we suggest possible future research directions and open questions in the point cloud registration field.
\end{itemize}

%\begin{table*}[t]  
%	\centering
%	\begin{tabular}{c|ccc } 
%		\hline
%		Type &Categories &  Sub-categories  & Publications\\
%		\hline\hline
%		%RL+BIC2 & 0 & 0 & 37 & 0 & 0 & 37 \\ 
%		%\hline
%		\multirow{8}{*}{Same-source}& \multirow{4}{*}{Optimisation-based registration methods} & ICP-based & \cite{bes1992method,yang2019polynomial,forstner2017efficient,segal2009generalized}\\
%		& & Graph-based & \cite{huang2017systematic,livi2013graph,duchenne2011tensor,zhu2019elastic}\\
%		& & GMM-based & \cite{huang2016coarse,huang2017coarse,myronenko2010point}\\
%		& & Semi-definite & \cite{le2019sdrsac,kezurer2015tight,dym2017ds++,Bernard_2018_CVPR,iglesias2020global}\\ \cline{2-4}
%		& \multirow{2}{*}{Feature-learning methods for registration} &Volumetric & \cite{3dmatch,gojcic2019perfect,choy2019fully,riegler2017octnet,wang2017cnn} \\
%		&&  Point Cloud & \cite{deng2018ppf,deng2018ppfnet} \\ \cline{2-4}
%		& \multirow{2}{*}{End-to-end learning-based registration}  & Regression & \cite{deng20193d,pais20193dregnet, lu2019deepvcp}   \\
%		& &Feature difference & \cite{aoki2019pointnetlk,huang2020feature,sarode2019pcrnet, wang2020alignnet}   \\
%		\hline
%		\multirow{2}{*}{Cross-source}& \multirow{2}{*}{Cross-source} &Optimization-based& \cite{huang2016coarse}\cite{huang2017coarse}\cite{huang2017systematic}\cite{huang2019fast}\cite{peng2014street}\\
%		& & {Learning-based} &\cite{huang2020feature}
%		\\\hline
%	\end{tabular}  
%	\caption{\textnormal{Taxonomy and representative publications of point cloud registration}}
%	\label{t_category}
%\end{table*}

\section{Problem definition}
Denote ${{\bf x}_i}^T (i\in[1,M])$ and ${{\bf y}_i}^T (j\in[1,N])$ as row vectors from matrices $X\in \mathbb{R}^{M\times 3}$ and $Y\in \mathbb{R}^{N\times 3}$ respectively. $X$ and $Y$ represent two point clouds, and ${\bf x}_i$ and  ${\bf y}_j$ are the coordinates of the $i_{th}$ and $i_{th}$ points in the point clouds respectively. Suppose $X$ and $Y$ have $K$ pairs of correspondences. The goal of registration is to find the rigid transformation parameters $g$ (rotation matrix $R \in \mathcal{SO}(3)$ and translation vector $t \in \mathbb{R}^3$) which best aligns the point cloud $X$ to $Y$ as shown below:
\begin{equation}
\begin{split}
% \operatorname*{arg\,min}_{R\in \mathcal{SO}(3), t\in \mathbb{R}^3} \sum_{i=1}^N r_i(F(P),F(RQ+t))  
\operatorname*{arg\,min}_{R\in \mathcal{SO}(3), t\in \mathbb{R}^3} \| d(X,g(Y)) \|_2^2 
% &=\sum_{i=1}^N r_i(P,RQ+t)
\end{split}
\label{formular} 
\end{equation}
where $d(X,g(Y)) = d(X, RY+t) =\sum\limits_{k=1}^{K}\|{\bf x}_k-(R{\bf{y}}_k+t)\|_2$ is the projection error between $X_k$ and transformed $Y_k$ $(k\in[1,K])$.  The equation \ref{formular} forms a well-known \textit{chicken-and-egg} problem: the optimal transformation matrix can be calculated if the true correspondences are known \cite{bes1992method}\cite{billings2015iterative}; in contrast, correspondences can also be readily found if the optimal transformation matrix is given. However, the joint problem cannot be trivially solved. The following sections are pieces of literature review about solving the registration problem.

\section{Challenges}
\label{challenges}
In this section, the same-source and cross-source point cloud registration challenges are summarized for both same-source and cross-source point cloud registration.
\subsection{Same-source challenges} As the point clouds are captured from the same type of sensors but different time or views, the challenges existed in the registration problem contain 
\begin{itemize}
	\item Noise and outliers. The environment and sensor noise are variant at different acquisition time, and the captured point clouds will contain noise and outliers around the same 3D position.
	\item Partial overlap. Due to different viewpoint and acquisition time, the captured point cloud is only partial overlapped.
\end{itemize}

\subsection{Cross-source challenges}
In recent years, point cloud acquisition has endured fast development. For instance, Kinect has been widely used in many fields. Lidar becomes use-affordable and has integrated into the mobile phone ( e.g. iPhone 12). Moreover, many years' development of 3D reconstruction has made the point cloud generation from RGB cameras possible. Despite these improvements in point cloud acquisition, each sensor contains its distinct advantages and limitations. For example, Kinect can record detailed structure information but has limited view distance; Lidar can record objects far away but has limited resolution. 
Many pieces of evidence \cite{peng2014street, huang2017systematic} show fused point clouds from different sensors could provide more information and generate better performance for applications. The point clouds fusion requires cross-source point cloud registration techniques.

Since the point clouds are captured from the different types of sensors, and different types of sensors contain different imaging mechanisms, the cross-source challenges in the registration problem are much more complicated than the same-source challenges. These challenges can be mainly divided into 
\begin{itemize}
	\item Noise and outliers. Because the acquisition environment, sensor noise and sensor image mechanisms are different at different acquisition time, the captured point clouds will contain noise and outliers around the same 3D position.
	\item Partial overlap. Due to different viewpoint and acquisition time, the captured point cloud is only partial overlapped.
	\item Density difference. Due to different imaging mechanisms and different resolutions, the captured point clouds usually contain different density.
	\item Scale variation. Since different imaging mechanisms may have different physical metrics, the captured point clouds may contain scale difference.
\end{itemize}

In this paper, we will conduct a comprehensive review of point cloud registration and build a new cross-source point cloud benchmark to evaluate the performance of the state-of-the-art registration methods in solving these challenges.

\section{Categories}
\begin{figure}
	\begin{subfigure}{.5\textwidth}
		\centering
		% include first image
		\includegraphics[width=.8\linewidth]{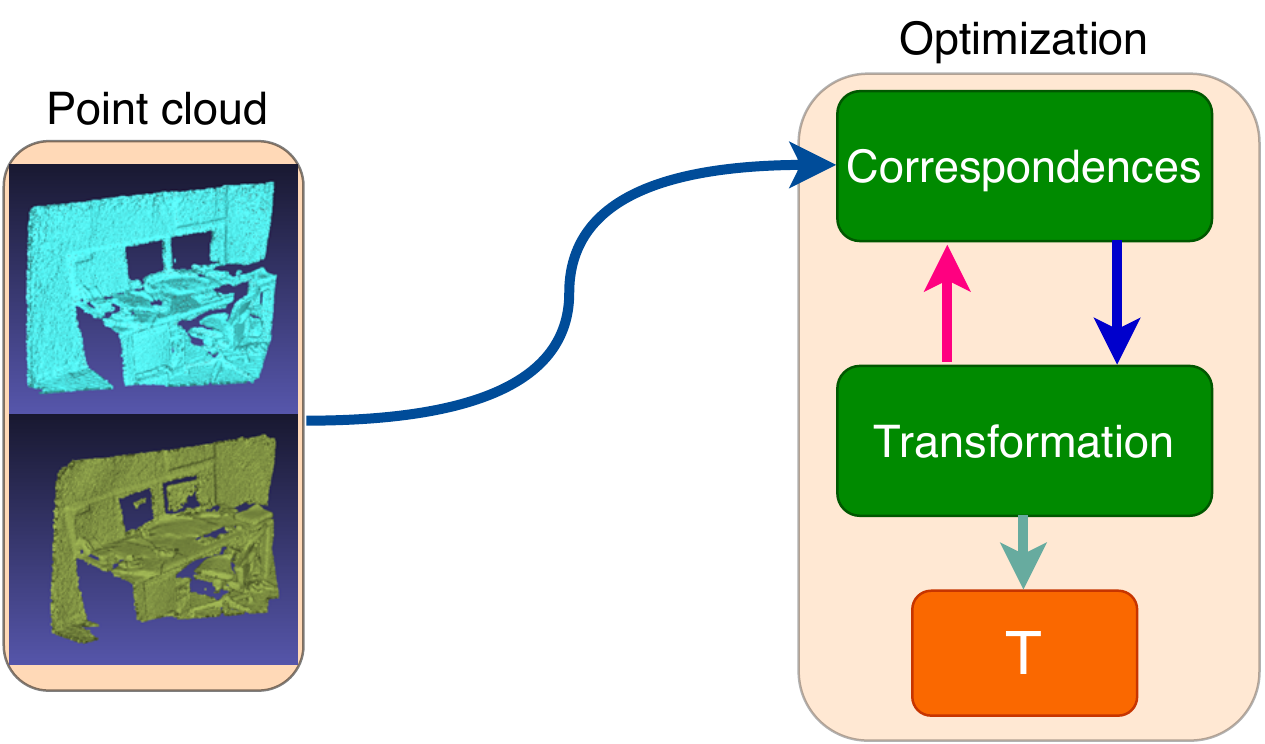}  
		\caption{An optimization-based framework for point cloud registration. Given two input point clouds, the correspondences and transformation between these point clouds are iteratively estimated. The algorithm outputs the optimal transformation \textbf{T} as the final solution. }
		\label{c1}
	\end{subfigure}
	\begin{subfigure}{.5\textwidth}
		\centering
		% include second image
		\includegraphics[width=.8\linewidth]{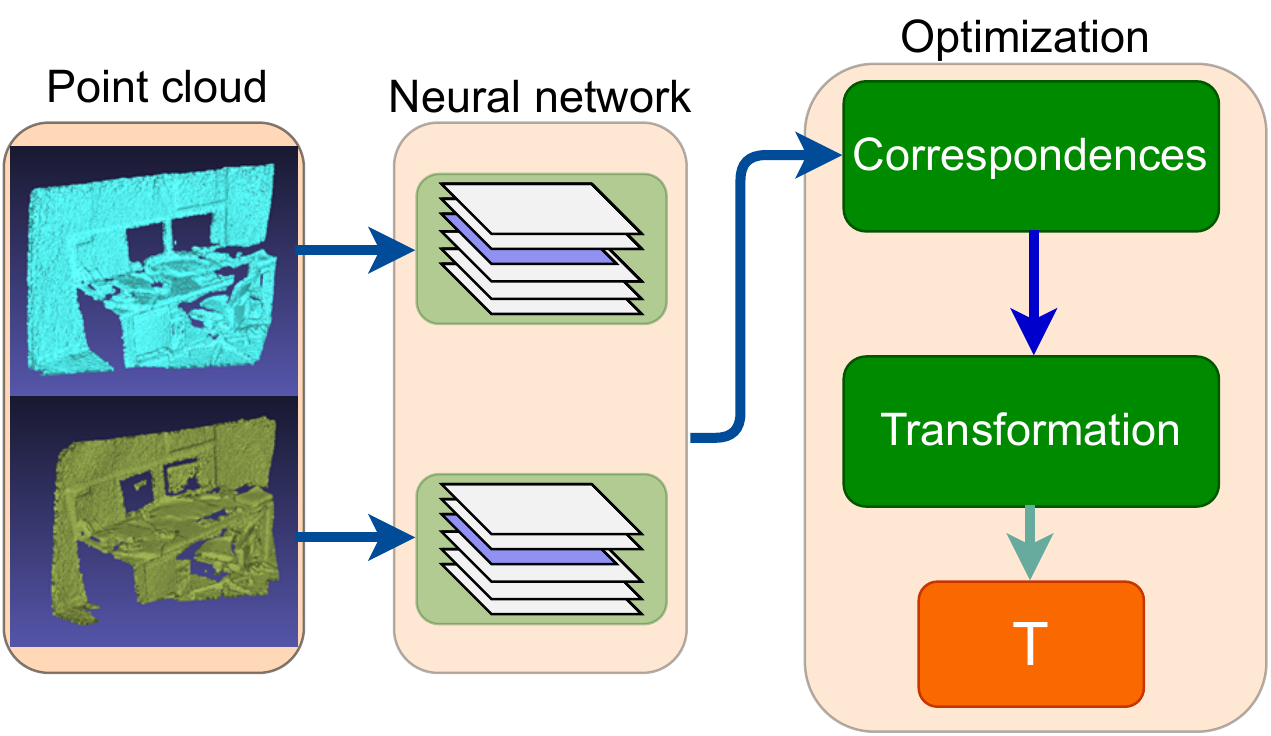}  
		\caption{A feature learning-based framework for point cloud registration. Given two input point clouds, the features are estimated using a deep neural network. Then, correspondence and transformation estimation run iteratively to estimate the final solution \textbf{T}.}
		\label{c2}
	\end{subfigure}	
	\begin{subfigure}{.5\textwidth}
		\centering
		% include third image
		\includegraphics[width=.8\linewidth]{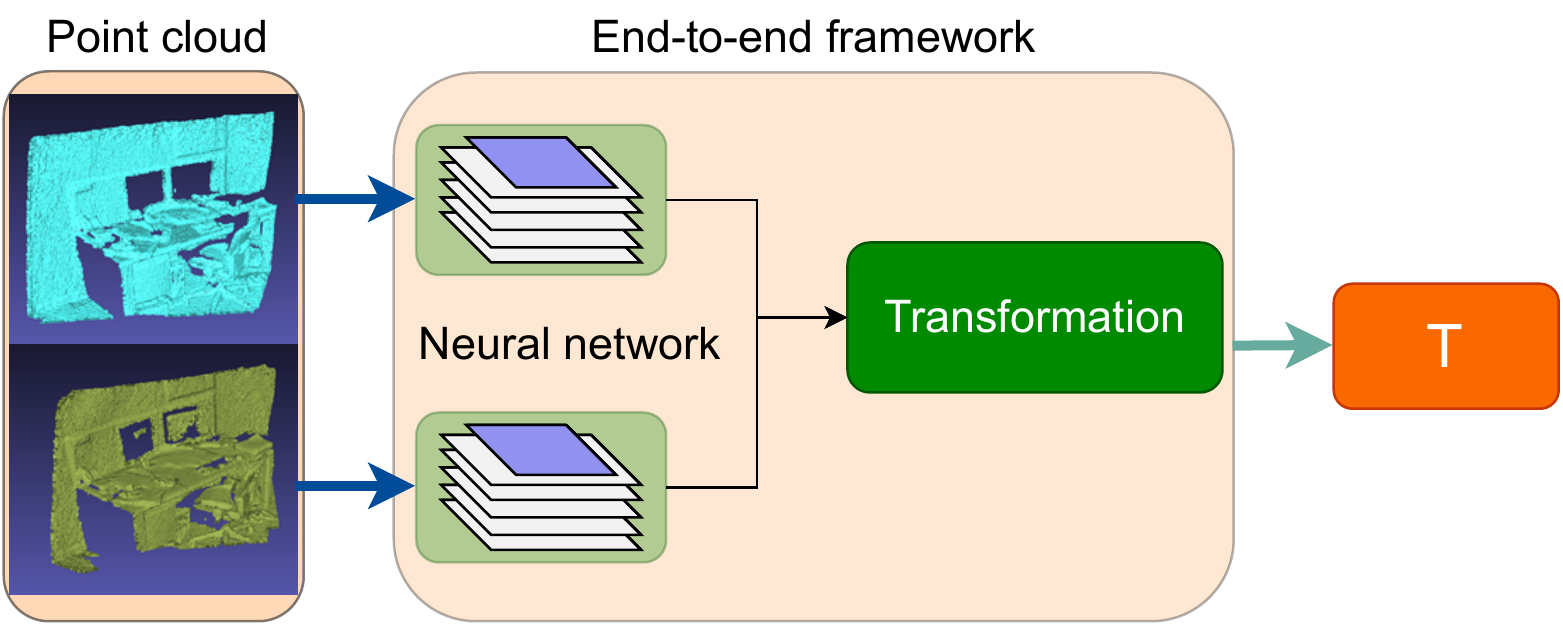}  
		\caption{An end-to-end learning-based framework for point cloud registration. Given two input point clouds, an end-to-end framework is used to estimate the final solution \textbf{T}.}
		\label{c3}
	\end{subfigure}
	\begin{subfigure}{.5\textwidth}
		\centering
		% include third image
		\includegraphics[width=.8\linewidth]{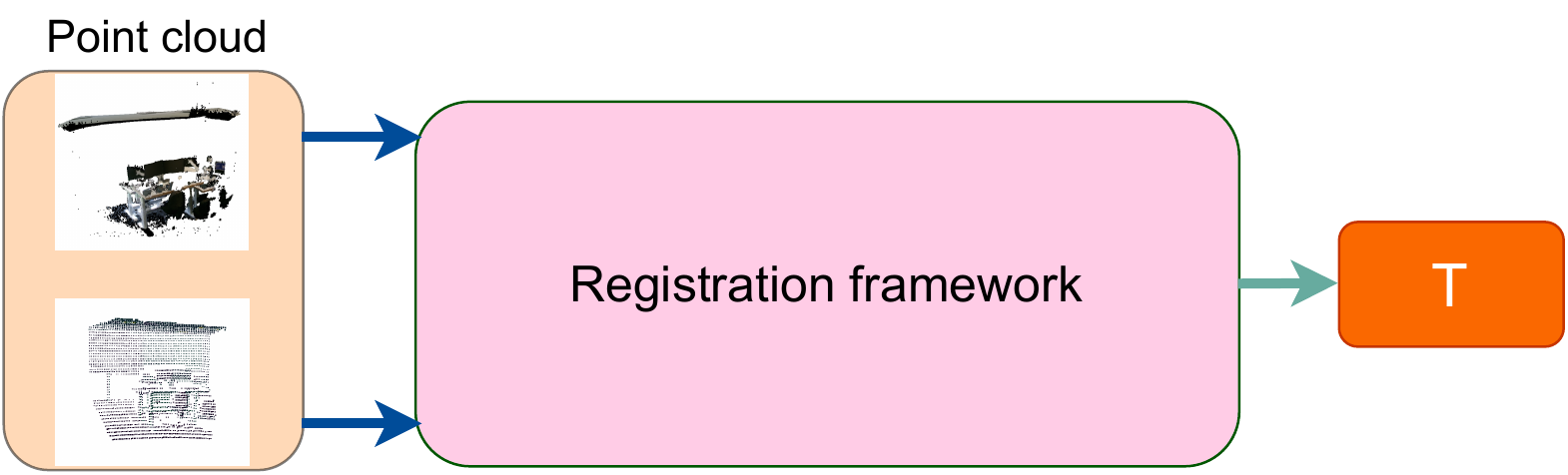}  
		\caption{An framework for cross-source point cloud registration. Given two input point clouds, a registration framework is designed to overcome cross-source challenges and estimate the final solution \textbf{T}.}
		\label{c4}
	\end{subfigure}
	\caption{Different frameworks to solve the same-source point cloud registration problem.}
	\label{category}
\end{figure}

This section presents our taxonomy of point cloud registration, as shown in Figure \ref{f_category}. We categorize point cloud registration into two types: same-source and cross-source registration. The same-source registration can be further divided into optimization-based registration methods, feature-learning methods, end-to-end learning registration. Figure \ref{category} summarizes the frameworks of these categories. In the following, we give a brief introduction to each category and analyze its advantages and limitations. 

\subsection{Optimisation-based registration methods}
Optimization-based registration is to use optimization strategies to estimate the transformation matrix. Most optimization-based registration methods \cite{yang2019polynomial,le2019sdrsac,pomerleau2015review,cheng2018registration} contain two stages: correspondence searching and transformation estimation. Figure (\ref{c1}) summarizes the main process of this category. Correspondence searching is to find the matched point for every point in another point clouds. Transformation estimation is to estimate the transformation matrix by using the correspondences. These two stages will conduct iteratively to find the optimal transformation. During the iterative process, the correspondences maybe not accurate at the beginning. The correspondences will become more and more accurate as the iterative process continues. Then, the estimated transformation matrix will become accurate by using precise correspondences. The correspondences can be found by comparing point-point coordinate difference or point-point feature difference.

The advantages of this category are two folds: 1) rigorous mathematical theories could guarantee their convergence. 2) They require no training data and generalize well to unknown scenes. The limitations of this category are that many sophisticated strategies are required to overcome the variations of noise, outliers, density variations and partial overlap, which will increase the computation cost.

\subsection{Feature learning methods for registration}
Unlike the classical optimization-based registration methods, feature learning methods \cite{3dmatch,deng2018ppfnet,gojcic2019perfect} use the deep neural network to learn a robust feature correspondence search. Then, the transformation matrix is finalized by one step estimation (e.g. RANSAC) without iteration. Figure (\ref{c2}) summarizes the primary processes of this category. For example, \cite{3dmatch} uses AlexNet to learn a 3D feature from an RGB-D dataset.  \cite{deng2018ppfnet} proposes a local PPF feature by using the distribution of neighbour points and then input into the network for deep feature learning. \cite{gojcic2019perfect} proposes a rotation-invariant hand-craft feature and input it into a deep neural network for feature learning. All these methods are using deep learning as a feature extraction tool. By developing sophisticated network architectures or loss functions, they aim to estimate robust correspondences by the learned distinctive feature. 

The advantages of this category are two folds: 1) deep learning-based point feature could provide robust and accurate correspondence searching. 2) The accurate correspondences could lead to accurate registration results by using a simple RANSAC method. The limitations of this kind of methods are three aspects: 1) they need large training data. 2) The registration performance drops dramatically in unknown scenes if the scenes have a large distribution difference to the training data. 3) They use a separated training process to learn a stand-alone feature extraction network. The learned feature network is to determine point-point matching other than registration.

\subsection{End-to-end learning-based registration methods}
The end-to-end learning-based methods solve the registration problem with an end-to-end neural network. The input of this category is two point clouds, and the output is the transformation matrix to align these two point clouds. The transformation estimation is embedded into the neural network optimization, which is different from the above feature-learning methods, whose focus is point feature learning. The neural network optimization is separate from the transformation estimation. Figure (\ref{c3}) summarizes the primary process of this category. The basic idea of end-to-end learning methods is to transform the registration problem into a regression problem. For example, \cite{Yang_2019_CVPR} tries to learn a feature from the point clouds to be aligned and then regresses the transformation parameters from the feature. \cite{wang2019non} proposes a registration network to formulate the correlation between source and target point sets and predict the transformation using the defined correlation. \cite{elbaz20173d} proposes an auto-encoder registration network for localization, which combines super points extraction and unsupervised feature learning. \cite{lu2019deepicp} proposes a keypoint detection method and estimates the relative pose simultaneously. FMR \cite{huang2020feature} proposes a feature-metric registration method, which converts the registration problem from the previous minimizing point-point projection error to minimizing feature difference. This method is a pioneer work of feature-metric registration by combining deep learning and the conventional Lucas-Kanade optimization method.

The advantages of this category are two folds: 1) the neural network specifically designs and optimizes for registration task. 2) It could leverage both the merits of conventional mathematical theories and deep neural networks. The limitations of current methods are two aspects: 1) the regression methods regard transformation parameter estimation as a black box, and the distance metric is measured in the coordinate-based Euclidean space, which is sensitive to noise and density difference. 2) the feature-metric registration method does consider the local structure information, which is very important for registration.  

\begin{figure*}
	\centering
	\includegraphics[width=\linewidth]{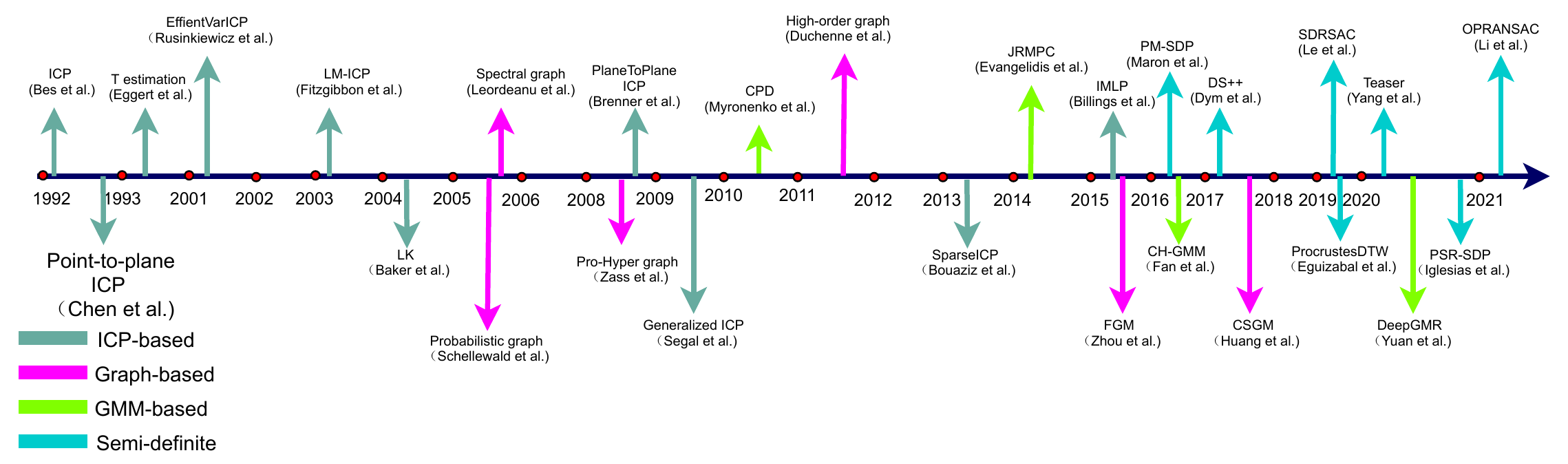}
	\caption{Chronological overview of the most relevant optimization-based methods.}
	\label{path_opt}
\end{figure*}
\subsection{Cross-source registration}
Cross-source point cloud registration is to align point clouds from different types of sensors, such as Kinect and Lidar. According to \cite{peng2014street,huang2017systematic}, cross-source point cloud registration is much more challenging because of the combination of considerable noise and outliers, density difference, partial overlap and scale difference.  Several algorithms \cite{huang2016coarse,huang2017systematic,huang2017coarse,huang2019fast} use sophisticated optimization strategies to solve the cross-source point cloud registration problem by overcoming the cross-source challenges. For example, CSGM \cite{huang2017systematic} transforms the registration problem into a graph matching problem and leverage the graph matching theory to overcome these challenges. Recently, FMR \cite{huang2020feature} shows performance on aligning cross-source point cloud using deep learning. These methods are trying hard to use optimization strategies or deep neural networks to estimate the transformation matrix by overcoming the cross-source challenges.

The benefit of cross-source point cloud registration is to combine several sensors' advantages and provide comprehensive 3D vision information for many computer vision tasks, such as augmented reality and building construction. However, the limitation is that the existing registration methods show low accuracy and high time complexity, which remain at infancy. With the recent fast development of 3D sensor technologies, the lack of cross-source point cloud registration research brings up a gap between sensor technology and cross-source applications.

From section \ref{optimizationMethod} to \ref{crosssourceMethod}, we contribute to summarize the key ideas and review the recent critical development of each category. In section \ref{connections}, we also summarize the connection between optimization-based methods and deep learning based on our literature review.

\section{Optimisation-based registration methods}
\label{optimizationMethod}
The critical ideas of optimization-based methods are to develop a sophisticated optimization strategy to achieve the optimal solution of the non-linear problem in Equation \ref{formular}. This non-linear problem becomes challenging because of the impact of same-source challenges (see Section \ref{challenges}). Figure (\ref{c1}) summarizes the primary process of this category. Based on the optimization strategy, this section presents an overview of four types of optimization methods: ICP-based variations, graph-based, GMM-based and semi-definite registration methods. Several milestone methods are illustrated in Fig. \ref{path_opt}.

\subsection{ICP-based registration}
ICP-based registration methods contain two main steps: correspondence estimation and transformation estimation. The critical research ideas are two parts, as shown in Figure \ref{c1}: robust correspondence estimation and accurate transformation estimation.

Correspondences are two points that localize in the same position of an object or scene, where each point comes from a different point cloud. Correspondence estimation becomes challenging with the impact of the above discussed same-source challenges. There are three types of distance metric: point-point, point-plane, and plane-plane metric to get correspondences. We will give details about these distance metrics and review the related literature. 

The point-point metric uses point-point coordinate distance or feature distance to find the closest point pair as a correspondence. Many variations following this concept are proposed to get better correspondences. For example, ICP \cite{bes1992method} uses the original point-point distance metric. EfficientVarICP \cite{rusinkiewicz2001efficient} summarizes the ICP process and proposes several strategies to improves the algorithm speed of the ICP process. IMLP \cite{billings2015iterative} improves the ICP by incorporating the measurement noise in the transformation estimation. 

Apart from the point-point distance metric, point-to-plane metric \cite{chen1992object, ramalingam2013theory, khoshelham2016closed} is to estimate the transformation parameters by minimizing the orthogonal distance between the points in one point cloud and the corresponding local planes in the other. Specifically, the point-to-plane algorithms run a similar way to point-point methods but minimize error along the surface normal, such as 
\begin{equation}
\begin{split}
\operatorname*{arg\,min}_{R\in \mathcal{SO}(3), t\in \mathbb{R}^3} \{\sum_{k=1}^K w_k\|{\bf n_k}*({\bf x}_k- (R{\bf y}_k+t)\|^2)\}
% &=\sum_{i=1}^N r_i(P,RQ+t)
\end{split}
\label{pplane} 
\end{equation}
where $w_k$ is the weights of each correspondence, $\bf n_k$ is the surface normal at point ${\bf x}_k$, ${\bf x}_k$ and ${\bf y}_k$ are point-correspondence pairs on point cloud $X$ and $Y$.

Segal et al. \cite{segal2009generalized} propose a generalized ICP to allows for the inclusion of arbitrary covariance matrices in both point-to-point and point-to-plane variants of ICP. The objective is to optimize
\begin{equation}
	\begin{split}
		\operatorname*{arg\,min}_{T} \{\sum_{k=1}^K \|d^T(C_k^Y + {\bf T}C_k^X{\bf T}^T)^{-1}d\|^2\}
		% &=\sum_{i=1}^N r_i(P,RQ+t)
	\end{split}
	\label{pgicp} 
\end{equation}
where $\{C_k^X\}$ and $\{C_k^Y\}$ are covariance matrices associated with the point cloud $X$ and $Y$. ${\bf T}$ is the transformation parameters that consists of $R$ and $t$, $d$ is a distance metric. For standard point-to-point ICP, it is a special case by setting $C_k^Y=I$ and $C_k^X=0$. Also, for point-to-plane ICP is a limiting case of this generalized ICP by setting $C_k^Y=P_k^{-1}$ and $C_k^X=0$, where $P_k^{-1}$ is the surface normal at $x_k$. The generalized ICP can also be applied to plane-to-plane ICP. The basic idea is to consider the point cloud is a sampled 2D manifold and use the local surface normal to represent the points.

In addition, plane-to-plane distance metric \cite{brenner2008coarse, khoshelham2010automated,forstner2017efficient} is adopted to estimate the correspondences. The objective is similar to point-point distance metric, which is 
\begin{equation}
\begin{split}
\operatorname*{arg\,min}_{R\in \mathcal{SO}(3), t\in \mathbb{R}^3} \{\sum_{k=1}^K \|{\bf nx}_k- (R{\bf ny}_k+t\|^2)\}
\end{split}
\label{p2p} 
\end{equation}
where $\bf nx$ and $\bf ny$ are surface normal of point cloud $X$ and $Y$. 

Regarding the transformation matrix, there are four kinds of methods: SVD-based \cite{bes1992method}, Lucas-Kanade (LK) algorithm \cite{baker2004lucas} and Procrustes analysis \cite{dryden2016statistical}. Given correspondences, the SVD-based estimation methods \cite{bes1992method,segal2009generalized, bouaziz2013sparse,yang2015go,billings2015iterative} perform singular value decomposition (SVD) to the difference of correspondences. Low et al.\cite{low2004linear} propose a linear approximation of the rotation matrix and estimate the transformation using SVD. It obtains much faster efficiency and more accuracy. LK algorithm \cite{baker2004lucas} estimates transformation using Jacobian of feature difference and approximation methods (e.g. Gauss-Newton). LM-ICP \cite{fitzgibbon2003robust} leverages the Levenberg-Marquardt algorithm to estimate the transformation by adding a damping factor to the original LK algorithm.  This method replaces the Euclidean distance with the Chamfer distance and uses a Levenberg-Marquardt algorithm to compute $T_k$. The  LM-ICP  method is superior to the standard ICP method, especially in treating high overlapping ratios. ICP \cite{bes1992method} proposes a closed-form solution by using singular value decomposition (SVD) to calculate the transformation matrix. Eggert et al. \cite{eggert1997estimating} summarise transformation estimation methods in four categories and compare their performance.

A Procrustes registration (rotation, scale, and translation,
as defined in \cite{dryden2016statistical}) converts the transformation estimation as a linear least-squares problem. The final pose ($P$) can be estimated as a closed-form solution $P=(X_2^HX_1)^{-1}X_2^Hx_1$, where $x_1$ and $x_2$ is the input point clouds, $X_2=[x_2 \ \ \bf 1]$. Since Procrustes registration requires given correspondences, the performance is highly relied on the accuracy of correspondence searching. ProcrustesDTW \cite{eguizabal2019procrustes} propose Dynamic Time Warping (DTW) \cite{muller2007information} to establish an automatic correspondence between the landmark-based shapes to be registered, which avoids the need for initial manual correspondence and same landmark-set lengths. This analysis is only conducted experiments on 2D, and further research on 3D is required.

\subsection{Graph-based registration}
Graph-based registration is another popular methods. The mains idea of graph-based registration is to tackle point cloud registration using a non-parametric model \cite{zhu2019review}. Since a graph consists of edges and vertexes, GM methods aim to find the point-point correspondences between two graphs by considering both vertexes and edges. This correspondence searching problem in GM methods can be considered as an optimization problem. The research direction is to develop a better graph matching optimization strategy to find more accurate correspondences. As shown in Figure \ref{c1}, accurate correspondences could contribute to a better transformation estimation.

To solve the optimization problem, based on objective functions' constraints, we can divide the GM methods into two categories: second-order methods and high-order methods. Second-order GM methods measure both the vertices-to-vertices and edges-to-edges similarity \cite{livi2013graph}. High-order GM methods involve more than two points, such as similarity of triangle pairs \cite{duchenne2011tensor}.

The optimization of graph matching belongs to the quadratic assignment problem (QAP) \cite{loiola2007survey}, which is an NP-hard problem \cite{garey1979computers}. The key to solving this QAP problem is to design approximation strategies. Based on their approximation method, we divide the second-order GM methods into three categories: doubly stochastic relaxation, spectral relaxation and semi-definite programming relaxation. Using a doubly stochastic matrix, the optimizing GM is transformed as a non-convex QAP problem. Therefore, many methods only find a local optimum. For example, \cite{almohamad1993linear} uses a linear program to approximate the quadratic cost. CSGM \cite{huang2017systematic} uses a linear program to solve the graph matching problem and apply it to solve the cross-source point cloud registration task. High-order graph \cite{duchenne2011tensor} uses an integer projection algorithm to optimize the objective function in the integer domain. FGM \cite{zhou2015factorized} factorizes the large pairwise affinity matrix into some smaller matrices. Then, the graph matching problem is solved with a simple path-following optimization algorithm. Spectral graph \cite{leordeanu2005spectral} uses a spectral relaxation method to approximate the QAP problem. The semi-definite programming (SDP) relaxation is to relax the non-convex constraint using a convex semi-definite. Then, a randomized algorithm \cite{torr2003solving} or a winner-take-all method \cite{schellewald2005probabilistic} is applied to find the correspondences between graphs.  

High-order graph matching methods is to compare the hyper-edges or hyper-nodes to find the correspondences. The advantage of high-order GM methods is that they are invariant to affine variations (e.g. scale difference). For example, Zass et al. \cite{zass2008probabilistic} design a probabilistic approach to solve the high-order graph matching problem. Duchenne et al. \cite{duchenne2011tensor} designs a triangle similarity and convert the graph matching problem into a tensor optimization problem. Recently, Zhu et al. \cite{zhu2019elastic} propose an elastic net to control the trade-off between the sparsity and the accuracy of the matching results by incorporating the Elastic-Net constraint into the tensor-based graph matching model. These methods are all affine-invariant.

\subsection{GMM-based registration }
Gaussian mixture models (GMM) is also a popular kind of methods in solving point cloud registration. The critical idea of GMM-based methods is to formulate the registration problem of Equation (\ref{formular}) into a likelihood maximization of input data. After the optimization, both the transformation matrix and parameters of Gaussian mixture models are calculated. The advantages of the GMM-based method are robust to noise and outliers \cite{bishop2006pattern, rasoulian2012group} since these methods align the distributions. The research direction is to develop an optimization strategy to optimize the transformation matrix by maximizing the likelihood.

CPD \cite{myronenko2010point}  introduces a motion drift idea into the  GMM  framework by adding constraints to transformation estimation. CH-GMM \cite{fan2016convex} combines the convex hull (a tighter set of original point set) and  GMM  to reduce the computation complexity.  JRMPC \cite{evangelidis2014generative} recasts the registration as a clustering problem, where the transformation is optimizing by solving the GMM. Recently, DeepGMR \cite{yuan2020deepgmr} uses deep learning to learn the correspondences between GMM components and points, and the transformation and GMM parameters can be estimated by a forward step.  

\subsection{Semi-definite registration} 
The main idea of semi-definite registration is to develop sophisticated approximation strategies. The reason is that the correspondences optimization of equation (\ref{formular}) is a quadratic assignment problem when considering paired correspondences constraint. Global optimization of such problem is an NP-hard problem \cite{loiola2007survey}. However, a good approximation to the global solution of correspondences can be achieved. If we define the correspondences assignment matrix as $W=\{0,1\}$, $w_{ij}=1$ means point $i$ is correspondent with point $j$ and 0 otherwise. The original correspondence assignment matrix is not semi-definite as the eigenvalue value $\lambda_{min}$ is not guaranteed to be non-negative. 

The research direction is to build different projection for the original correspondences so that the estimation of $W$ can be a semi-definite optimization problem.  This subsection describes several popular ways to convert the original optimization of the equation (\ref{formular}) into a semi-definite optimization problem. 

\textbf{Symmetric matrix.} To estimate the correspondences, we introduce a symmetric matrix $A \in \mathbf{R}^{N^2\times N^2}$ describes the matching potentials between pairs of points and $Y=\|X\|^2$. The eigenvalue of $Y$ is non-negative. According to SDRSAC \cite{le2019sdrsac} and DS++ \cite{dym2017ds++}, the optimization of correspondences is to solve the problem of $\max_{XY} AY$ with four conditions: (1) $X$ should be $\{0,1\}$, (2) the row sum of $X$ should be no larger than 1, (3) the column sum of $X$ should be no larger than 1, and (4) the sum of $X$ should equal to the number of correspondence pairs. By solving the above maximization problem, we can obtain the global solution of correspondences. The transformation can be calculated with a closed-form solution by using the correspondences \cite{huang2017coarse}. Recently, there are several algorithms \cite{enqvist2009optimal, le2017alternating, leordeanu2005spectral} focus on solving non-rigid registration. They have all shared a similar theory of semi-definite relaxation.

\textbf{Laplacian matrix.} The point cloud registration problem in Equation \ref{formular} can also be re-written in more compact by using trace notation as follows:

\begin{eqnarray}
\begin{split}
S=\min_{R,t} \sum_{i,j} \| y_i - (R_jx_{ij}+t_j)\|^2 \\
 = tr(YLY^T-2YX^TR^T)
\end{split}
\end{eqnarray}

where $L=A-WB^{-1}W^T$ is the Laplacian of a weighted graph, each corresponding to a 3D point from $Y$. $A_{ii}=\sum_{j}w_{ij}$ and $B_{jj}=\sum_{i}w_{ij}$ are diagonal matrices, $w_{ij}=\{0,1\}$ determines if point $i$ of $X$ is matched with point $j$ of $Y$. Since the graph Laplacian has positive semi-definite properties, this problem can be solved using semi-definite relaxation. Recently, PSR-SDP\cite{iglesias2020global} uses the semi-definite relaxation to solve the multiple point sets registration. Teaser\cite{yang2020teaser} uses graduated non-convexity to solve the rotation sub-problem. This strategy leverages Douglas-Rachford Splitting to certify global optimality efficiently. This method solves the high computation cost in SDP relaxation. Recent OPRASANC \cite{li2021point} introduces a graduated optimization strategy to largely alleviate the effect of local minima and obtains better efficiency than Teaser.

Semi-definite relaxation is a strong convex relaxation that achieves the global minimum of the original problem. However, semi-definite relaxation usually faces the scalability problem \cite{kezurer2015tight} which is only tractable for up to 15 points. \cite{kolmogorov2006convergent} utilizes the Markov random field techniques to approximate their linear programming relaxation solution. PM-SDP\cite{maron2016point} obtains better efficiency by reducing the dimension of semi-definite constraints. However, they still can only handle the middle size of the point cloud registration. The efficiency is still a remaining research problem.

\begin{figure*}
	\centering
	\includegraphics[width=\linewidth]{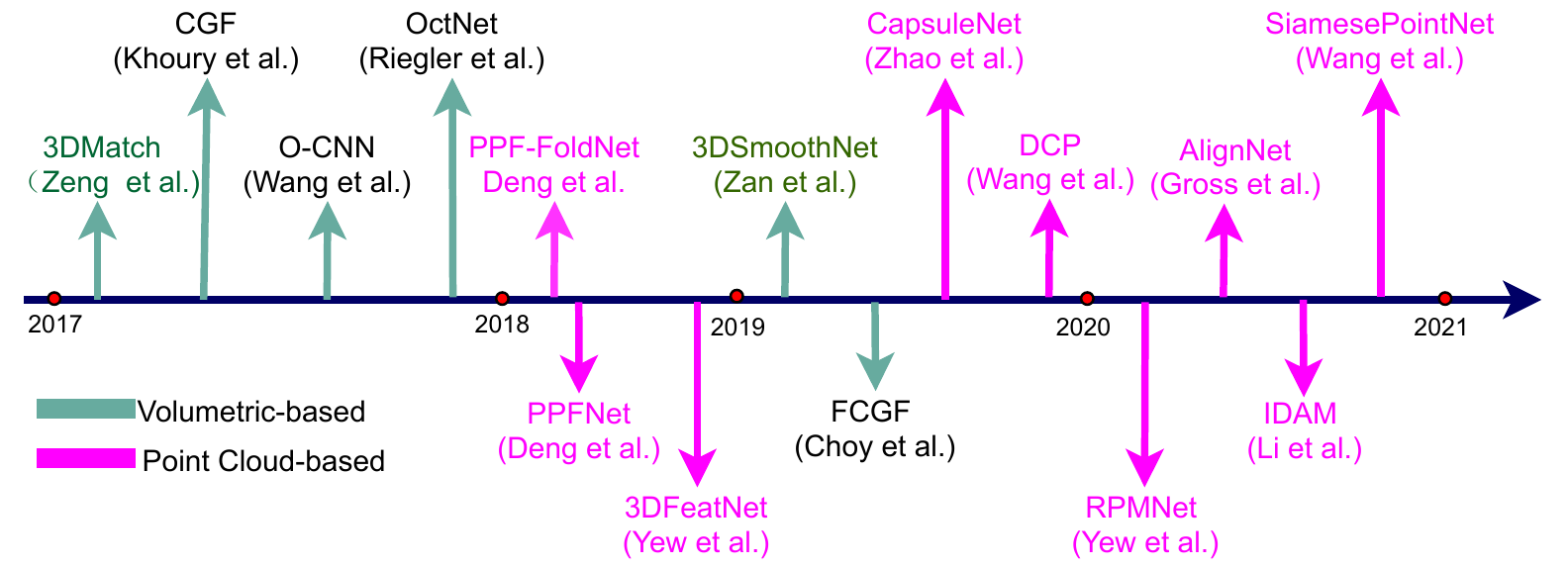}
	\caption{Chronological overview of the most relevant feature-learning registration methods.}
	\label{path_feature_learning}
\end{figure*}   
\section{Feature-learning methods for registration}
The main idea of feature-learning methods is to use the deep feature to estimate accurate correspondences. Then, the transformation can be estimated using one-step optimization (e.g. SVD or RANSAC) without iteration between correspondence estimation and transformation estimation, as shown in Figure \ref{c2}. The research direction is to design advanced neural networks to extract distinctive features. In this section, several feature-learning registration methods are reviewed. Regarding the data format of deep learning, these registration methods are divided into learning on volumetric data and point cloud. Several milestone methods are illustrated in Fig. \ref{path_feature_learning}.

\subsection{Learning on volumetric data} 
\begin{figure}[h]
	\includegraphics[width=\linewidth]{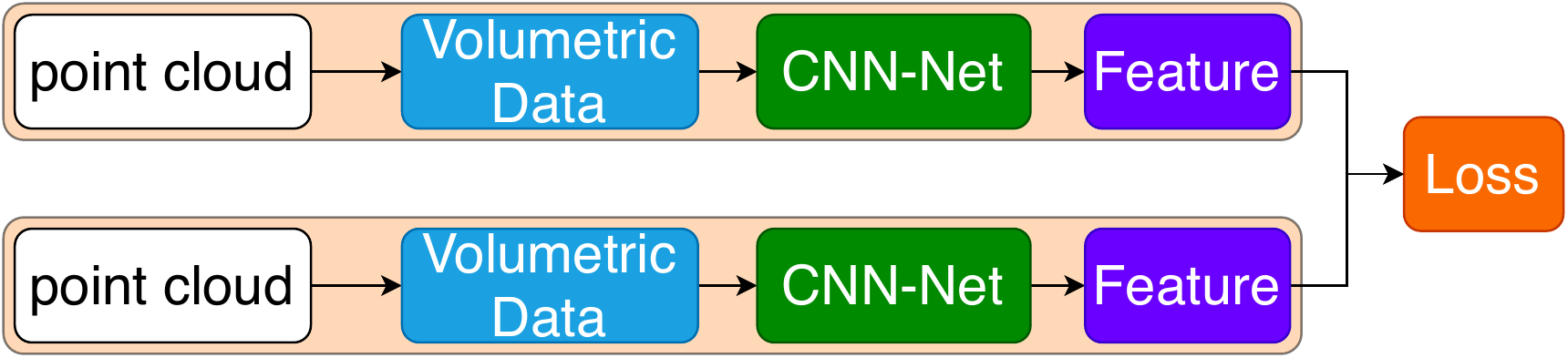}
	\caption{The overall framework of 3DMatch, which is an example of neural networks in Figure \ref{c2} using volumetric data.}
	\label{f3dmatch}
\end{figure}

3DMatch \cite{3dmatch} trains a parallel network from RGBD images. The input of 3DMatch is 3D volumetric data, and the output is a 512-dimensional feature for a local patch. 3DMatch can extract a local feature for 3D point clouds. Figure \ref{f3dmatch}  shows its overall framework, which is an example case of the neural network in Figure \ref{c2}. For each interest point of a 3D point cloud, the 3DMatch is to extract a feature to incorporate the local structure around the interest point. In 3DMatch, the 3D point cloud needs to convert into 3D volumetric data and then extract the local representation by feeding the 3D volumetric data into the neural network. This method has two obvious drawbacks: volumetric data requires large Graphic process unit (GPU)  memory and sensitive to rotation variations. 

3DSmoothNet \cite{gojcic2019perfect} introduces a pre-processing method to align the 3D patches and calculate the volumetric data based on the aligned 3D patches. By feeding the aligned volumetric data into a convolution neural network, the extracted features are rotation-invariant. Specifically, a local reference frame (LRF) is estimated using the eigendecomposition of all points' covariance matrix. After the point clouds are aligned using the LRF, Gaussian smoothing is applied to the input grids to get a  smooth density value (SDV) voxelization. Then, the SDV is fed into a network for feature extraction. To improve the efficiency of volumetric-based descriptor, FCGF \cite{choy2019fully} uses $1\times 1 \times 1$ kernel to extract a  fast and compact metric features for geometric correspondence. 

There is much literature that focuses on handling the limitation of large memory cost. The key idea is to remove empty voxels since the 3D point cloud is usually sparsely located in the 3D volumetric data. OctNet \cite{riegler2017octnet} uses Octree to hierarchically divide the volumetric data into an unbalanced tree where each leaf node stores the feature presentation. Tatarchenko et al. \cite{tatarchenko2017octree} use Octree to decode the point cloud and learns distinctive representation. Similarly, O-CNN \cite{wang2017cnn} proposes an octree-based convolution neural network for 3D shape analysis.

\subsection{Learning on point cloud}
\begin{figure}[h]
	\includegraphics[width=\linewidth]{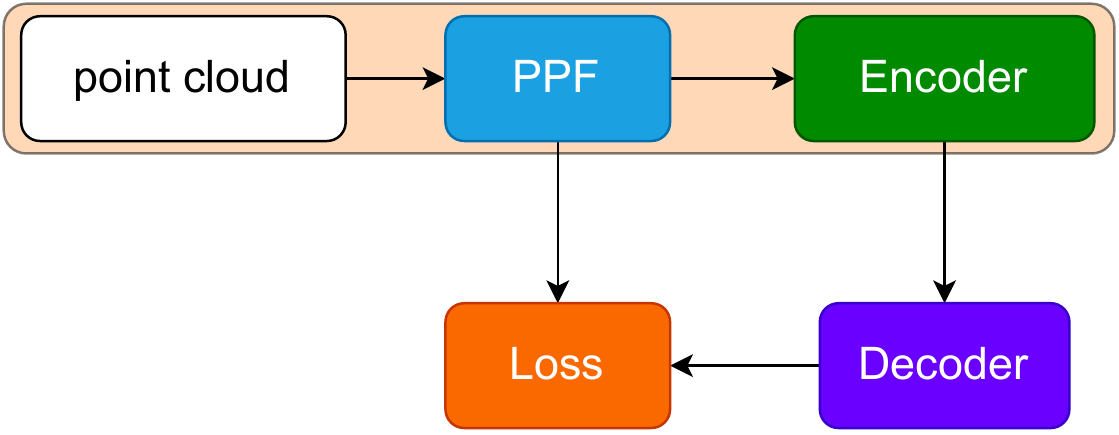}
	\caption{The overall framework of PPFNet,, which is an example of neural networks in Figure \ref{c2} using point cloud.}
	\label{fppffoldnet}
\end{figure}

Instead of feeding the network with volumetric data,  PPFNet \cite{deng2018ppfnet} learns local descriptors on pure geometry and is highly aware of the global context. This method uses a point pair feature (PPF) \cite{drost2010model} to pre-process the input point cloud patches to achieve rotation invariant. Then, the point clouds are input into a PointNet \cite{qi2017pointnet} to extract a local feature. Then, a global feature is obtained by applying a max-pooling operation. Both the global and local features are input in an MLP block to generate the final correspondence search feature. The limitation is that it requires a large amount of annotation data. To solve this issue, PPF-FoldNet \cite{deng2018ppf} proposes an unsupervised method to remove the annotation requirement constraint. The overall framework is shown in Figure \ref{fppffoldnet}. The basic idea is to use PointNet to encode a feature and use a decoder to decode the feature into data be the same as the input. The whole network is optimized by using the difference between the input and output using Chamfer loss. Similarly, SiamesePointNet\cite{zhou2020siamesepointnet} produces the descriptor of interest points by a hierarchical encoder-decoder architecture.

By not requiring manual annotation of matching point cluster, 3DFeatNet \cite{yew20183dfeat} introduces a weakly-supervised approach that leverages alignment and attention mechanisms to learn feature correspondences from GPS/INS tagged 3D point clouds without explicitly specifying them. More specifically, the network takes a set of triplets containing an anchor, positive and negative point cloud. They train the neural network with the triplet loss by minimizing the difference between the anchor and positive point clouds while maximizing the difference between the anchor and negative point clouds. Alignment \cite{gross2019alignnet} focuses on the partially observed object alignment by using a tracking framework, which is trying to estimate the object-centric relative motion. Moreover, this approach uses a neural network that takes the noisy 3D point segments of objects as input to estimate their motion instead of approximating targets with their centre points. \cite{yang2020color} utilizes both the colour and spatial geometric information to solve the point cloud registration. 

Since the ICP requires hard assignments of closest points, it is sensitive to the initial transformation and noisy/outliers. Therefore, the ICP usually converges to the wrong local minima. RPMNet \cite{Yew_2020_CVPR} introduces a less sensitive to initialization and more robust deep learning-based approach for rigid point cloud registration. This method's network can get a soft assignment of point correspondences and can solve the point cloud partial visibility.
The deep closest point (DCP) \cite{wang2019deep} employs a dynamic graph convolutional neural network for feature extraction and an attention module to generate a new embedding that considers the relationships between two point clouds. Besides, a singular value decomposition module is used to calculate rotation and translation. IDAM \cite{li2019iterative} incorporates both geometric and distance features into the iterative matching process. Point matching involves computing a similarity score based on the entire concatenated features of the two points of interest.
Yang et al. \cite{yang2020learning} find that more compact and distinctive representations can be achieved by optimizing a neural network (NN) model under the triplet framework that non-linearly fuses local geometric features in Euclidean spaces. The NN model is trained by an improved triplet loss function that fully leverages all pairwise relationships within the triplet. Moreover, they claimed that their fused descriptor is also competitive to deeply learned descriptors from raw data while being more lightweight and rotational invariant.

\section{End-to-end learning-based registration}
\begin{figure}[h]
	\centering
	\includegraphics[width=\linewidth]{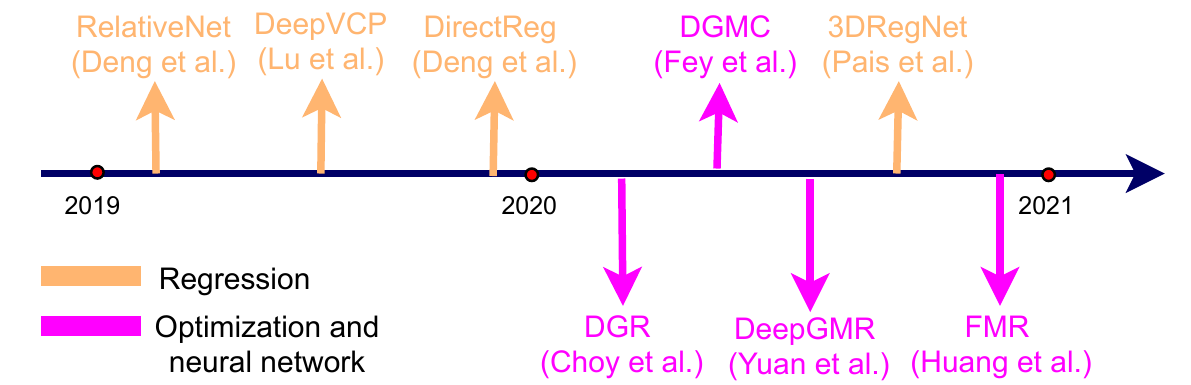}
	\caption{Chronological overview of the most relevant end-to-end learning registration methods.}
	\label{path_end_end}
\end{figure}

The main idea of end-to-end learning-base registration methods is that two-point clouds fed into the neural network, and output is the transformation matrix between these two point clouds. There are two categories: (1) considering the registration as a regression problem and using the neural network to fit a regression model for the transformation matrix estimation \cite{wang2019non, Yang_2019_CVPR, deng20193d,pais20193dregnet}; Figure \ref{end2end} shows the overall framework for these methods. (2) considering the registration as an end-to-end framework by the combination of neural network and optimization \cite{huang2020feature, choy2020deep}. Figure \ref{c3} shows the overall framework of these methods. These two categories aim to train a deep neural network to directly solve the registration problem in equation \ref{formular}. Several milestone methods are illustrated in Fig. \ref{path_end_end}.

\subsection{Registration by regression}
\begin{figure}[h]
	\includegraphics[width=\linewidth]{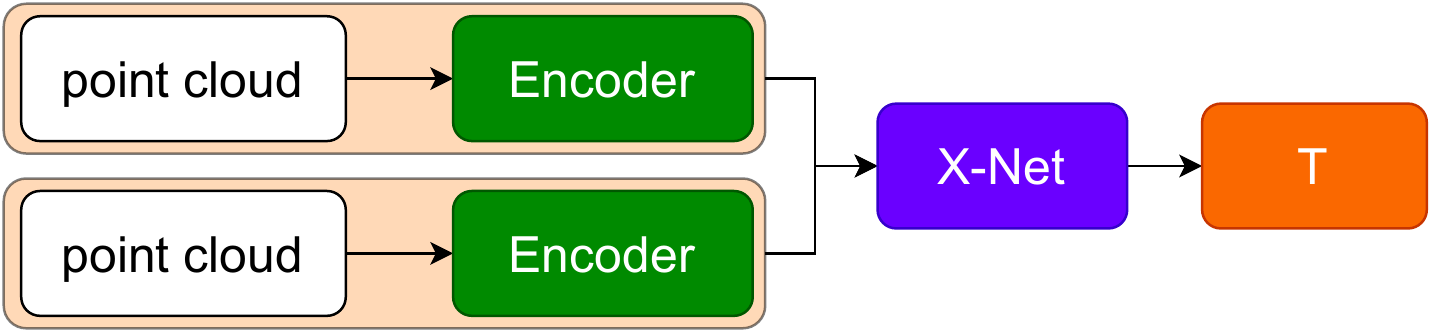}
	\caption{The overall framework of end-to-end learning-based regression methods, which is an example of Figure \ref{c3}. Two global features are extracted firstly. Then, the two features fed into a X-Net to estimate a transformation matrix $T$.}
	\label{end2end}
\end{figure}

Deng et al. \cite{deng20193d} propose a relativeNet to estimate the pose directly from features. Lu et al. \cite{lu2019deepvcp} propose a method (DeepVCP) to detect keypoints based on learned matching probabilities among a group of candidates, which can boost the registration accuracy. Pais et al. \cite{pais20193dregnet} develop a classification network to identify the inliers/outliers and uses a regression network to estimate the transformation matrix from the inliers. Figure \ref{end2end} shows the overall framework of these registration methods by regression. The connection to Fig. \ref{c3} is that the transformation module is implemented with an X-Net module.

\subsection{Registration by optimization and neural network}
The main idea of this category is to combine the conventional registration-related optimization theories with deep neural networks to solve the registration problem in Equation \ref{formular}. Figure \ref{c3} shows a summary of these methods. PointNetLK \cite{aoki2019pointnetlk} uses the PointNet\cite{qi2017pointnet} to extract global features for two input point clouds and then use a inverse compositional (IC) algorithm to estimate the transformation matrix. By estimating the transformation matrix, the objective is to minimize the feature difference between the two features. For this feature-based IC algorithm, the Jacobian estimation is challenging. PointnetLK uses an approximation method through a finite difference gradient computation. This approach allows the application of the computationally efficient inverse compositional Lucas-Kanade algorithm. 
Huang et al. \cite{huang2020feature} further improve PointNetLK with an autoencoder and a point distance loss. Meantime, it can reduce the dependence on labels. DeepGMR \cite{yuan2020deepgmr} uses a neural network to learn pose-invariant point-to-distribution parameter correspondences. Then, these correspondences are fed into the GMM optimization module to estimate the transformation matrix. DGR \cite{choy2020deep} proposes a 6-dimensional convolutional network architecture for inlier likelihood prediction and estimate the transformation by a weighted Procrustes module. These methods show that the combination of conventional optimization methods and recent deep learning strategies obtain better accuracy than previous methods.

%\begin{figure}[h]
%	\includegraphics[width=\linewidth]{optreg.pdf}
%	\caption{The overall framework of registration by optimization and neural network, which is an example of Figure \ref{c3}. Two global features are extracted firstly. Then, the two features fed into an optimization module to estimate a transformation matrix T.}
%	\label{optnet}
%\end{figure}

%\begin{table*}[t]
%	\centering
%	\begin{tabular}{ccc } 
%		\hline
%		Method & Framework&  Github link\\
%		3DMatch & marvin & https://github.com/andyzeng/3dmatch-toolbox\\
%		FMR & pytorch & https://github.com/XiaoshuiHuang/fmr\\
%		\hline\hline
%		%RL+BIC2 & 0 & 0 & 37 & 0 & 0 & 37 \\ 
%		%\hline
%
%		\hline
%	\end{tabular}
%	\caption{Summary of  Open-source Implementations.}
%	\label{t1}
%\end{table*}

\begin{table*}[t]
	\centering
	\begin{tabular}{ccccccccccc } 
		\hline
		Dataset &  sensor& sceneNum & indoor &outdoor & dense &sparse & ground-truth & xyz & corlor \\
		\hline\hline
		%RL+BIC2 & 0 & 0 & 37 & 0 & 0 & 37 \\ 
		%\hline
		3DMatch & depth & 56 & $\checkmark$ & $\times$ & $\checkmark$ & ${\times }$ & synthetic & $\checkmark$ & $\checkmark$ \\
		KITTI & LiDAR & 8 & $\times$ & $\checkmark$ & $\times$ & ${\checkmark }$ & synthetic & $\checkmark$ & $\times$ \\
		ETHdata & LiDAR & 8 & $\times$ & $\checkmark$ & $\times$ & ${\checkmark }$ & synthetic & $\checkmark$ & $\checkmark$ \\
		3DCSR &Indoor &21 & $\checkmark$  & $\checkmark$ &$\checkmark$ &$\checkmark$ & manual&$\checkmark$ & $\checkmark$ \\
		\hline
	\end{tabular}
	\caption{Summary of existing same-source and cross-source dataset.}
	\label{t1}
\end{table*}

\section{Cross-source point cloud registration}
\label{crosssourceMethod}
\begin{figure}[h]
	\includegraphics[width=\linewidth]{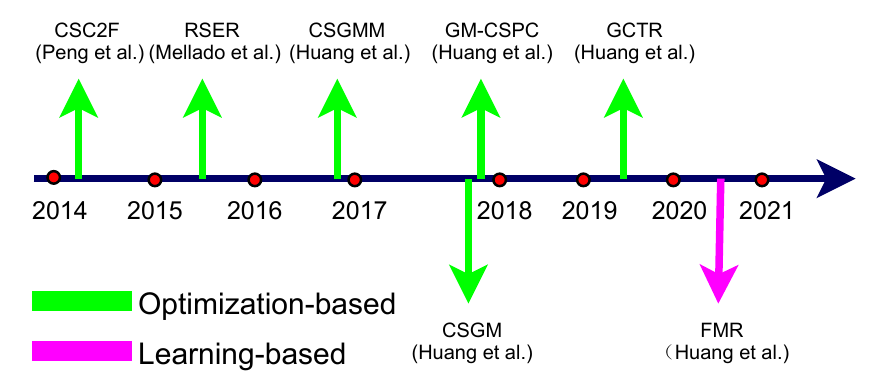}
	\caption{Chronological overview of the most relevant cross-source point cloud registration methods.}
	\label{path_cs}
\end{figure}

For the first time, a comprehensive review of cross-source point cloud registration is conducted in this section. The existing cross-source registration methods are divided into two categories: optimization-based methods and learning-based methods. The research direction is to design advanced registration framework (e.g. Fig. \ref{c4}) to overcome the cross-source challenges (discussed in section \ref{challenges}) and solve the Equation \ref{formular}.  Several milestone methods are illustrated in Fig. \ref{path_cs}.

\subsection{Optimization-based methods:} 
The main idea of optimization-based methods is to design sophisticated optimization strategies to solve the point cloud registration problem in the equation \ref{formular}. The optimization strategies are similar to the same-source registration but require a more complicated version to overcome the severe cross-source challenges. Since the registration algorithm is usually more complicated than the same source, the proposed algorithms are usually a registration framework. Figure \ref{c4} visually summarizes the ideas. CSC2F \cite{peng2014street} proposes a first cross-source point cloud registration method by using a coarse-to-fine method. The registration is solved by using ICP. Following the coarse-to-fine strategy, CSGMM \cite{huang2016coarse} applies GMM-based algorithm to estimate the transformation. GM-CSPC \cite{huang2017coarse} assumes the cross-source point clouds are coming from the same Gaussian mixture models and the two input point clouds are two samples from the Gaussian mixture. The GM-CSPC estimates both the GMM parameters and transformation simultaneously.  CSGM \cite{huang2017systematic} converts the registration problem into a graph matching problem and estimate the transformation matrix by graph matching optimization. Recently, \cite{huang2019fast} introduce high-order constraints to correspondences searching and convert the registration problem into a tensor optimization problem. RSER \cite{mellado2015relative} proposes a scale estimation method and use RANSAC to calculate the transformation after scale normalization.

The advantages of this category are the same as the same-source optimization-based registration methods, which contain are two folds. Firstly,  rigorous mathematical theories could guarantee their convergence or performance. Secondly, they require no training data and generalize well to unknown scenes. However, the challenges of this category are that the sophisticated strategies require large computation cost, and the performance of these methods is varying at different datasets.

\subsection{Learning-based methods:} 
Based on our knowledge, FMR \cite{huang2020feature} is the first learning-based method to solve the cross-source point cloud registration. This method combines the optimization and deep neural network and estimates the transformation by minimizing the global feature difference. This method has demonstrated considerable noise, outliers and density difference. Because the deep neural network is good at robust feature extraction, the learning-based method is a promising direction to solve cross-source point cloud registration.

Although there are many learning-based registration algorithms, the performance on the cross-source dataset is less reported. In this paper, we build a new cross-source point cloud benchmark and evaluate several state-of-the-art registration algorithms' performance on this benchmark. This comparison will provide some insights for future research.

\section{Connections between optimization-based methods and deep learning:} 
\label{connections}
The connections between deep learning and optimization-based methods are: the deep learning technique could serve as a feature extraction tool to replace the original point coordinate. The conventional optimization could provide a theoretical guarantee for the convergence. Firstly, advanced loss calculation strategies are developed to apply an optimization strategy to calculate an estimated transformation from the learned feature. Secondly, calculate the loss between the estimated transformation and ground truth. Many existing methods \cite{wang2019deep, huang2020feature} demonstrate that combining both advantages could achieve both high accuracy and efficiency. For instance, deep closest point (DCP)\cite{wang2019deep} uses deep features to estimate correspondences and use SVD to calculate the transformation.  FMR \cite{huang2020feature} applies deep learning to extract global feature and uses Lukas-Kanade (LK) algorithm to minimize the feature difference. Fey, M. et al. \cite{Feyetal2020} uses deep learning to calculate the soft correspondences and use message passing network to refine the correspondences. DeepGMR \cite{yuan2020deepgmr} uses deep learning to calculate the correspondences between Gaussian models and points and optimize the transformation based on GMM optimization. 

These existing approaches provide some initial trials on conventional optimization and deep neural networks to solve registration problems. However, both the accuracy robustness and efficiency are still required to improve further. Combining conventional optimization theory and recent deep neural networks is a promising way to provide high accuracy and efficiency and theoretically guarantee current deep learning-based registration methods. The research direction is to design advanced loss calculation strategies to optimize the neural network by combining the existing optimization strategies.

\section{Evaluations}
This section summarises the existing metrics and summarises the performance of existing methods on the existing same-source datasets. Then, we introduce a new cross-source dataset and conduct comparison experiments for the existing registration methods. This section will provide a benchmark for both same-source and cross-source point cloud registration.

\subsection{Evaluation metrics}
\textbf{rmseP:} Root square mean error of projection {(rmseP)} is calculated as the mean of point-point projection error after applying the transformation. 
\textbf{rmseT:} Root square mean error of transformation {(rmseT)} represents the root-mean-square error between estimated transformation $g_{est}$ and ground truth transformation $g_{gt}$.
\textbf{RE:} The rotation error (RE) is calculated as the Euclidean distance of rotation parameters between estimated $r_{est}$ and ground truth $r_{gt}$. The rotation parameters are angles on three axes.
\textbf{TE:} The translation error (TE) is calculated as the Euclidean distance of translation parameters between estimated $t_{est}$ and ground truth $t_{gt}$.  
\textbf{Recall:} The recall represents the number of point cloud pair that RE and TE are below a threshold to the total pair number. Alternatively, the rmseP is below a threshold.

\subsection{Same-source dataset}
\textbf{ModelNet40}
The ModelNet40 \cite{wu20153d} is a comprehensive clean collection of 3D CAD models for objects containing 40 categories and 13356 models in total. The CAD models of each category have divided into test and train parts. Each model contains several nodes and faces.  A random rotation and translation transform each model to evaluate the registration. The transformed model and the original model are utilized to evaluate the performance of the registration algorithm. 
%\begin{table}[h]
%	\centering
%	\begin{tabular}{ccc} 
%		\hline
%		Methods &  rotation error & translation error  \\
%		\hline\hline
%		FGR \cite{zhou2016fast} & 0.2 \\
%		RANSAC \cite{rusu2009fast}& 34.2\\
%		FCGF \cite{choy2019fully}& 98.2 \\
%		DGR \cite{choy2020deep}& 98.0\\
%		FGR \cite{zhou2016fast} & 0.2 \\
%		RANSAC \cite{rusu2009fast}& 34.2\\
%		FCGF \cite{choy2019fully}& 98.2 \\
%		DGR \cite{choy2020deep}& 98.0\\
%		FCGF \cite{choy2019fully}& 98.2 \\
%		DGR \cite{choy2020deep}& 98.0\\
%		\hline
%	\end{tabular}
%	\caption{Comparison on ModelNet40 datasets.}
%	\label{modelnet}
%\end{table}

\textbf{3DMatch}: 
contains a total of over 200K RGB-D images of 62 different scenes, such as  7-Scenes, SUN3D, RGB-D Scenes v.2 and Halber. Each scene is divided into several fragments. Each fragment is reconstructed from 50 depth frames using TSDF volumetric fusion and saved to a .ply file. The reconstruction datasets are captured in different environments with different local geometries at varying scales and built with different reconstruction algorithms. During the experiments, fifty-four scenes are used for training and eight scenes for testing.  
\begin{table}[h]
	\centering
	\begin{tabular}{ccc} 
		\hline
		Methods &  Average Recall &Thresholds  \\
		\hline\hline
		ICP(p2point)\cite{zhou2018open3d}& 6.04 & TE(0.3m),RE(15$^\circ$)\\
		ICP(p2plane)\cite{zhou2018open3d}& 6.59 & TE(0.3m),RE(15$^\circ$)\\
		Super4PCS\cite{mellado2014super}& 21.6 & TE(0.3m),RE(15$^\circ$)\\
		GO-ICP\cite{yang2015go}& 22.9 & TE(0.3m),RE(15$^\circ$)\\
		FGR\cite{zhou2016fast}& 42.7 & TE(0.3m),RE(15$^\circ$)\\
		RANSAC \cite{rusu2009fast} & 66.1 & TE(0.3m),RE(15$^\circ$)\\
		\hline
		SpinImage \cite{johnson1999using} & 34 & rmseP($0.2m$)\\
		SHOT \cite{salti2014shot} & 27 & rmseP($0.2m$) \\
		FPFH \cite{rusu2009fast} & 40 & rmseP($0.2m$ \\
		USC \cite{tombari2010unique}& 43 & rmseP($0.2m$)\\
		PointNet \cite{qi2017pointnet}& 48 & rmseP(0.2m)\\
		CGF \cite{khoury2017learning} & 56 & rmseP($0.2m$)\\
		3DMatch \cite{3dmatch} & 67 & rmseP($0.2m$)\\ 
		PPFNet \cite{deng2018ppfnet} & 71 & rmseP($0.2m$)\\
		FCGF \cite{choy2019fully}& 82 & rmseP($0.2m$) \\
		DGR \cite{choy2020deep} & 91.3 & TE(0.3m),RE(15$^\circ$)\\
		\hline
		PointNetLK \cite{aoki2019pointnetlk}& 1.61 & TE(0.3m),RE(15$^\circ$)\\
		DCP \cite{wang2019deep}& 3.22 & TE(0.3m),RE(15$^\circ$)\\
		\hline
	\end{tabular}
	\caption{Comparison on 3DMatch datasets.}
	\label{t3dmatch}
\end{table}

\textbf{KITTI}:
The odometry dataset is initially designed for stereo matching performance evaluation, which contains stereo sequences, Lidar point clouds, and ground truth poses. It consists of 22 stereo sequences, where 11 sequences (00-10) have ground-truth trajectories for training, and 11 sequences (11-21) have no ground truth for evaluation. The Lidar point clouds are captured by using a Velodyne laser scanner.

\begin{table}[h]
	\centering
	\begin{tabular}{ccc} 
		\hline
		Methods &  Average Recall &Thresholds \\
		\hline\hline
		FGR \cite{zhou2016fast} & 0.2 &  TE(0.6m),RE(5$^\circ$) \\
		RANSAC \cite{rusu2009fast}& 34.2&  TE(0.6m),RE(5$^\circ$) \\
		FCGF \cite{choy2019fully}& 98.2 &  TE(0.6m),RE(5$^\circ$) \\
		DGR \cite{choy2020deep}& 98.0&  TE(0.6m),RE(5$^\circ$) \\
		FPFH \cite{rusu2009fast} & 58.95 &TE(2m),RE(5$^\circ$)  \\
		USC \cite{tombari2010unique}& 78.24 & TE(2m),RE(5$^\circ$) \\
		CGF \cite{khoury2017learning} & 87.81 & TE(2m),RE(5$^\circ$) \\
		3DMatch \cite{3dmatch} & 83.94 & TE(2m),RE(5$^\circ$) \\		
		3DFeatNet \cite{yew20183dfeat} & 95.97 & TE(2m),RE(5$^\circ$) \\
		\hline
	\end{tabular}
	\caption{Comparison on KITTI datasets.}
	\label{tkitti}
\end{table}

\textbf{ETHdata}
This group of datasets was recorded with Laser, IMU and GPS sensors. The point clouds are captured by using Hokuyo UTM-30LX. A theodolite is utilized to guarantee the precision of the "ground truth" positions of the scanner be in the millimetre range. The dataset contains eight scenes which consist of two indoor, five outdoor and one mixed environment.  Each scene contains around 30 fragments and stores them in a CSV file. The dataset contains global aligned frames and local frames with ground-truth transformation.
\begin{table}[h]
	\centering
	\begin{tabular}{ccc} 
		\hline
		Methods &  Average Recall & Thresholds \\
		\hline\hline
		FPFH \cite{rusu2009fast} & 67 &TE(2m),RE(5$^\circ$)  \\
		USC \cite{tombari2010unique}& 100 & TE(2m),RE(5$^\circ$) \\
		CGF \cite{khoury2017learning} & 92.1 & TE(2m),RE(5$^\circ$) \\
		3DMatch \cite{3dmatch} & 33.3 & TE(2m),RE(5$^\circ$) \\
		3DFeatNet \cite{yew20183dfeat}& 95.2&TE(2m),RE(5$^\circ$) \\
		
		\hline
	\end{tabular}
	\caption{Comparison on ETHdata datasets.}
	\label{tETHdata}
\end{table}

\subsection{New cross-source benchmark}

The above literature review shows that most of the current research focuses on same-source point cloud registration. While several existing methods \cite{peng2014street,huang2016coarse,mellado2015relative, huang2019fast} are targeted on cross-source domains, the accuracy is low and time complexity is huge, which remain at infancy. There is a gap between sensor technology and cross-source applications. We believe this is a large part attributed to the lack of an appropriate dataset. 

In this paper, we introduce a benchmark dataset for cross-source point cloud registration to bridge this gap. Specifically, the dataset is captured using recent popular sensors: LiDAR, Kinect and camera sensors. In total, 202 pairs of point clouds, where two scenes are captured using Kinect and RGB camera, 19 scenes are acquired from LiDAR and Kinect sensors. The dataset contains the most common objects or scenes in an indoor workspace environment. We manually align them to obtain the ground-truth transformation. Because different types of sensors have different imaging mechanisms and sensor noise, their acquired cross-source point clouds mainly contain cross-source challenges, as discussed in Section \ref{challenges}. The proposed dataset could serve as a dataset to evaluate the point cloud registration on cross-source data.

\subsubsection{Benchmark dataset: 3DCSR}
We have two kinds of cross-source point cloud: (1)- Kinect and Lidar and (2) Kinect and 3D reconstruction.

\textbf{Kinect and Lidar:} Kinect and Lidar active sensors are used to capture the same scene separately. For Kinect data, we use the KinectFusion \cite{whelan2012kintinuous} in Microsoft SDK 2.0 to generates point clouds. For Lidar data, we use the integrated software to record. During the acquisition, Lidar is stable in a scene for one capture and Kinect record several parts in this Lidar captured scene. For each scene, it is a point cloud reflect an indoor workspace. There are 19 scenes in this category. For each scene, there are different Kinect point clouds.  In total, we generate 165 pairs of cross-source point clouds using Kinect and Lidar. 

\textbf{Kinect and 3D reconstruction:} Kinect and iPhone RGB camera are utilized to record the same object/scene separately. The Kinect-based point cloud is generated using KinectFusion, and the camera-based point cloud is generated using VSFM \cite{wu2011visualsfm} from 2D images. The recorded point clouds are cross-source point clouds. There are two scenes in this acquisition: point clouds of 18 simple indoor objects and 19 multiple objects. In total, we generate 37 pairs of cross-source point clouds using Kinect and iPhone camera.

\textbf{Annotation:}
Before annotating, we manually remove the apparent outliers in the air of KinectFusion and 3D reconstructed point clouds. Then, we crop a related part from the Lidar scene. After then, we sample the Kinect data to 3 million for simple scenes and 4 million for complex scenes with the \textit{CloudCompare} software. When the sampled Kinect point cloud, 3D reconstructed point cloud and cropped Lidar point cloud are ready, we manually align them to get the annotations. For each pair, we cost one computer science expert and cross-check with two other experts. 

\begin{figure}[t]
	\includegraphics[width=\linewidth]{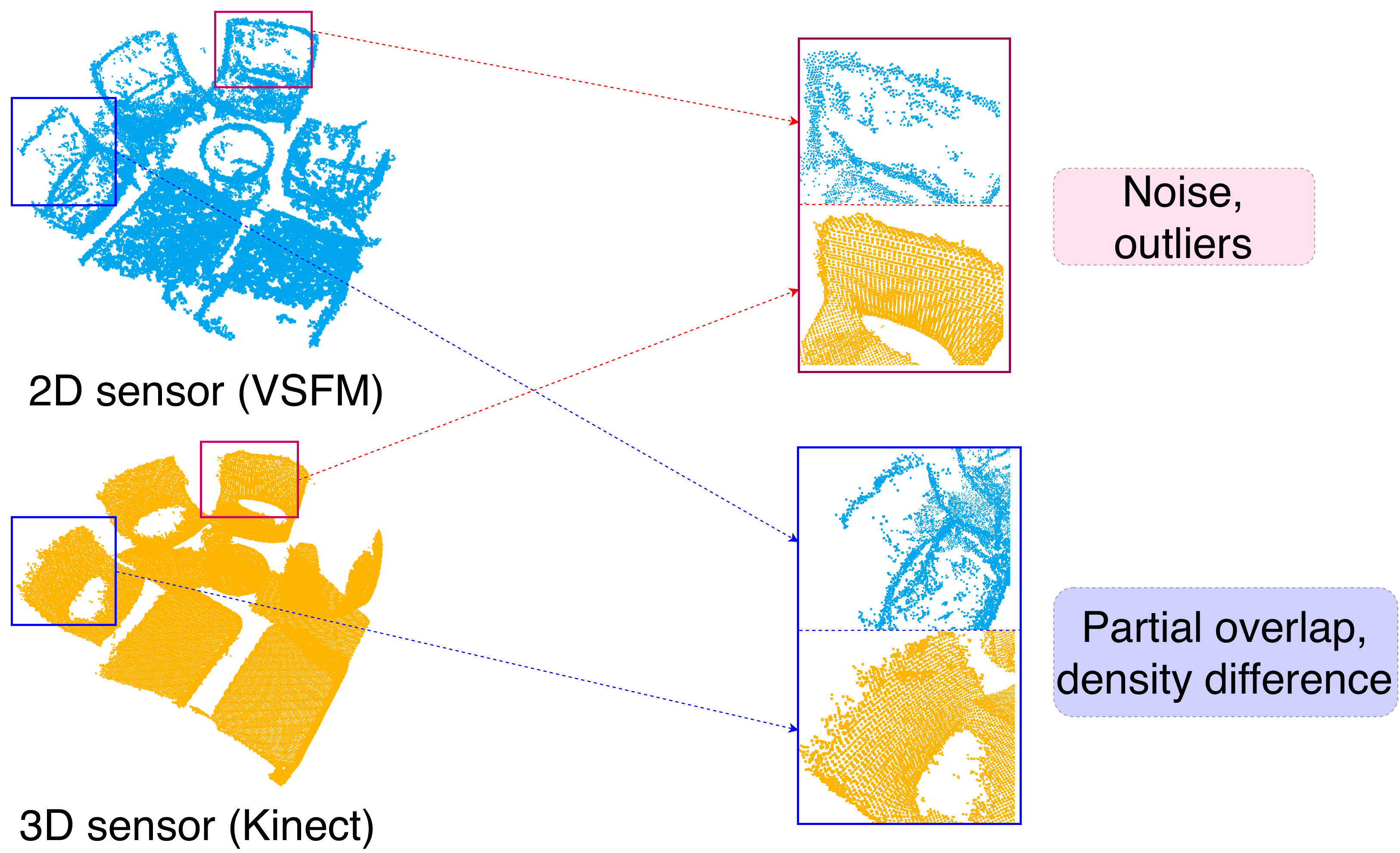}
	\caption{An example shows the challenges of cross-source point clouds. Considerable noise, outlier, density difference and partial overlap universally exist in the cross-source pair.}
	\label{f1}
\end{figure}

\textbf{Challenges:}
There is a mixture of variations of noise, outliers, density difference, and partial overlap for cross-source point clouds. See section \ref{challenges} for detailed explanation.  Figure \ref{f1} shows an example to demonstrate the challenges in cross-source point clouds.

\subsubsection{Evaluation}
Then, we run evaluation experiments for two objectives: (1) evaluating the state-of-the-art point cloud registration algorithms on the proposed benchmark dataset; (2) providing a research direction based on their performance. The registration recall is calculated as the number of point cloud pair that $RE<15^\circ$ and $TE<0.3$m to the total pair number.

{Baselines:}
%In this paper, the main contributions are a novel registration method and a new overlap region estimation approach.
(1) \textbf{Same-source point cloud registration.}
{FGR} \cite{zhou2016fast} is selected for the classic optimization-based algorithm. The FPFH descriptor is used for FGR. FMR \cite{huang2020feature} represents the feature-metric registration, which uses a semi-supervised approach to optimize a feature-metric error.  DGR \cite{choy2020deep} is a representative for the correspondence-learning registration that uses feature learning to get correspondences and integrates with a weighted Procrustes algorithm. DGR has already demonstrated better performance than the state-of-the-art feature-learning methods. Both FMR and DRG are trained in 3DMatch and evaluated on the proposed cross-source benchmark.

(2) \textbf{Cross-source point cloud registration.} Since \cite{peng2014street} uses ICP, \cite{huang2016coarse} uses Gaussian mixture model alignment, \cite{mellado2015relative} uses RANSAC to solve the cross-source registration problem, we only compare their registration parts. We also re-implement and compare with GCTR \cite{huang2019fast}, which is a recent work focus on cross-source point cloud registration. Due to the huge memory cost of Gaussian mixture model and huge computation cost of GCTR, we follow their original papers to uniform sample the original point clouds to approximately 2000 and 200 for GMM alignment \cite{huang2016coarse} and GCTR \cite{huang2019fast} respectively.

\begin{table}[h]
	\begin{center}
		%\begin{tabular}{p{1.5cm}p{1cm}p{1cm}p{1cm}p{1cm}p{1cm}p{1cm}}
		\begin{tabular}{p{1.0cm}|p{1.2cm}|p{0.8cm}|p{0.8cm}|p{1.0cm}|p{1.0cm}}
			\hline
			Type& Method & Recall & TE & RE(deg)  & Time(s)\\
			\hline
			\multirow{3}{*}{\parbox{1cm}{Same  source}}
			&FGR          &1.49\%  &0.07  & 10.74&2.23\\
			&PointnetLK    &0.50\%  &0.09  & 12.54 &2.25\\
			&FMR          &17.8\% &0.10  &4.66  &0.28\\
			&DGR 	     &36.6\% &0.04  & 4.26 &0.87\\ 
			%&DCP 	     &36.6\% &0.04  & 4.26 &0.87\\ 
			%&FCGF 	     &36.6\% &0.04  & 4.26 &0.87\\ 
			%&TEASER 	     &36.6\% &0.04  & 4.26 &0.87\\ 
			%&DeepGMR 	     &36.6\% &0.04  & 4.26 &0.87\\ 
			\hline
			\multirow{5}{*}{\parbox{1cm}{Cross  source}}
			&\cite{peng2014street} &24.3\% &0.38  &5.71  &0.19\\
			&\cite{huang2016coarse}  &1.0\% &0.71  &8.57  &18.1\\
			&\cite{mellado2015relative} &3.47\% &0.13  &8.30  &0.03\\
			&GCTR\cite{huang2019fast}   &0.50\% &0.17  &7.46  &15.8\\
			%&CSReg       &\bf 66.8\% &\bf 0.031  & \bf3.35 & 0.41\\
			\hline
		\end{tabular}
	\end{center}
	\caption{Quantitative comparisons on the cross-source dataset.}
	\label{crosssource}
\end{table}

Table \ref{crosssource} shows that the current state-of-the-art registration algorithms, including optimization-based (FGR), feature-metric (FMR) and correspondence learning (DGR) methods, are still facing difficulty to align cross-source point clouds. Among these existing methods, DGR obtains the best performance in solving the cross-source point cloud registration problem. 

Since cross-source point clouds contain cross-source challenges, keypoint-based methods may be a promising research direction. The reason lies in that robust keypoint extraction could find the key information from the noisy point clouds and overcome the cross-source challenges. For example, the DGR uses neural networks to generate a high probability for critical correspondences. Then, these key correspondences could play a critical role in transformation estimation by using a weighted Procrustes algorithm. That is the reason for the best performance.

\section{Applications}
\label{applications}
Point cloud registration is a critical technique in many applications. This section introduces the point cloud registration role in various applications and summarizes the research directions in each application.
\subsection{Construction}

BIM (Building Information Modelling) is a new generation of information storage and manipulation systems that widely used for construction purpose and building management. It usually contains  3D model and properties of the building. Previous computer-aided BIM design are limited to simple guides and theoretical planning since there is no interact with the real physical world. 

\begin{figure}[h]
	\includegraphics[width=\linewidth]{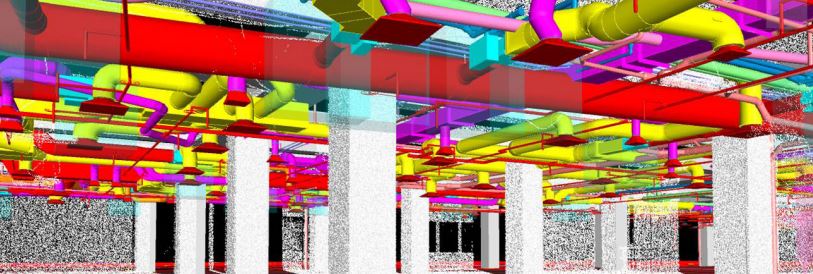}
	\caption{Point cloud in BIM model \cite{hayes2015use}.}
	\label{construct}
\end{figure}

Point cloud can overcome this limitation and offer the ability to align the digital models with physical space in exacting detail (see Figure \ref{construct} as an example). The reason is that point cloud provides the ability to effectively import 3D physical space into a digital format and augment your existing digit models. Point cloud will make dynamic evaluation, visualization and renovation projects much easier.

Although point cloud will bring technology renovation in construction \cite{chen2013point}, two obstacles limit its wide applications.  Firstly, 3D sensors are costly. A Leica RTC360 Laser Scanner Kit could cost about \$100,000. Secondly, the efficiency is low ($>20$ minutes to capture a $360^\circ$ scene). The main factor of low efficiency lies in the slow registration algorithm. Although some improvements are proposed for point cloud registration \cite{kim2018automated, kim2017automatic, zhang2012robust,liu2020deformation}, there is still a lack of robust and fast registration algorithms. The requirement of construction is high precision. Developing a fast and high accurate registration algorithm with construction field knowledge is urgent and will contribute to the construction field.

\subsection{Mining space}
In the mining area, the point cloud can provide a 3D experience mine and aid in monitoring underground tunnel wall movement and detecting pit wall instability, confirming development heading, and various other applications. For example, drone surveys and underground scanning equipment are changing how mining companies see their mine, giving them the ability to access a nearly real-time view of the terrain and development progress. The point cloud registration is the fundamental technology that dominates the success of these applications.

Point cloud has become a key data component for planning, operations and decision-making in the mining fields \cite{rodriguez2017underground}. For example, \cite{monsalve2018preliminary, monsalve2019application} conclude that the integration of terrestrial laser scanning (TLS) with discrete element modelling (DEM) can be used to prevent rock falls in underground excavations to enhance worker safety, which will reduce the fatality rate. However, it requires an adequate rock mass characterization and structural mapping where point cloud registration is the key technology. \cite{lee2019analyzing} uses point cloud to measure the vertical safety pillar volume and analyze the stability of the underground mine environment. The point cloud can also be used to build the terrain, which provides benefit for the survey of mining regions \cite{zhu2018lidar}.

%Recently, laser sensor combined with Unmanned Aerial Vehicles (UAVs) provides a renovation in the mining area. Imagine a mine where everyday planning software automatically tasks a fleet of UAVs – completely autonomously – to collect high-resolution coordinate scans, imagery and other remote sensing of the entire mine. Data from highwalls, stockpiles, waste dumps, tailings dams, blasting, and plants is collected by the same software and converted into information for quicker, smarter decision making. This scenario describes a future that will soon be reality. UAVs are already having a profound effect on mining.  For example, there have long been many places in a mine where foot traffic is not allowed or is ill-advised. These include near the crests and toes of highwalls, under operating machinery, on stockpiles and muck piles, and near blasts. Under these circumstances, obtaining measurements with a surveying rod, total station or GNSS is problematic. UAV aerial photography and remote sensing allow us to capture all that information without putting someone in harm's way. 

\begin{figure}[h]
	\includegraphics[width=\linewidth]{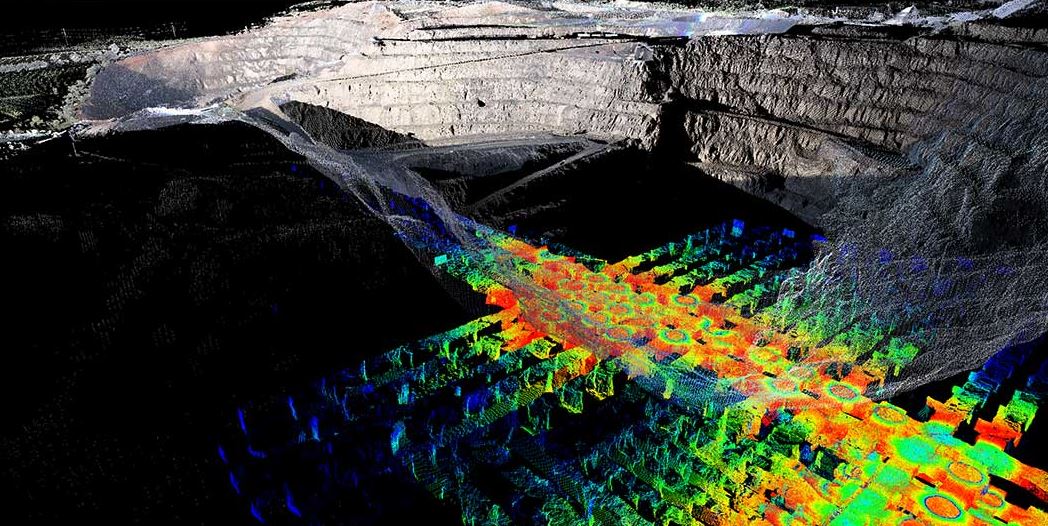}
	\caption{Coal mine volume estimation.} 
	\label{mining}
\end{figure}

The above applications show that point cloud brings a lot of great ideas in mining areas. However, all these applications require high-quality point clouds. Registration of point cloud is the key technology to merge multiple scans to a single larger scan. The registration accuracy will dominate the quality of these applications. For example, we cannot obtain high-quality coal mine volume estimation without accurate registration (See Figure \ref{mining} \footnote{A picture from https://www.maptek.com/products/pointstudio/index.html}). Developing highly accurate and fast point cloud registration with mining field knowledge will contribute the mining industry.

\subsection{Autonomous driving}
Recently, 3D sensors have widely applied in autonomous driving, which provides highly accurate 3D environment sensing data. The point cloud is an efficient way to store these 3D data. Since each sensor has view limitation in each scan, point cloud registration is crucial to provide high-quality 3D data with a larger view for autonomous driving. The main contribution of registration includes two aspects: create a larger 3D scan and provide pose estimation.

A High-resolution 3D map provides autonomous driving eyes, the critical data for navigation, planning, and localization. Construction of such map requires the registration algorithm \cite{poto2017laser, ilci2020high}. The quality of the registration algorithm dominates the quality of the high-resolution 3D map.
Moreover, point cloud registration between real-time point cloud of a vehicle and 3D map can apply for real-time vehicle localization \cite{nagy2018real, pan2019clustermap}. There is a review of 3D point cloud processing and learning for autonomous driving \cite{chen20203d}.

The key requirement of autonomous driving is high accuracy and real-time efficiency if used for localization. Developing high accurate and fast registration algorithms with prior road information is the research direction in autonomous driving.  
\subsection{Robotics}
When the 3d sensor is implanted on a robotic, the point cloud registration can be used to generate a 3D map to support firefighters and first responders in search and rescue missions. Those missions include nuclear incident, chemical spill and some other dangerous situations. Also, the point cloud registration with robotics can be used for Power Plant Inspection \cite{tache2009magnebike}. The robotics can also be used to monitor the shoreline \cite{aguilar20173d}. For example, \cite{hitz2016state} developed an autonomous surface vessel with a 3D laser to support freshwater bodies' environmental monitoring. Recently, the UAV with a laser sensor can also be used to do the survey \cite{turner2016uavs}. Recently, \cite{pomerleau2015review} proposes a review of point cloud registration in robotics. Among these applications, point cloud registration is the key technology. Accuracy and efficiency are the key requirements for the registration algorithms. Proposing fast and accurate registration with robotic field knowledge is very urgent and has high value for robotics fields.

\subsection{Other applications}
The point cloud is now an indispensable geological and geotechnical data for geomechanical analysis \cite{lyons2016applications}. Since point cloud acquisition is efficient, using the point cloud technologies such as registration could easily compare the difference between the point cloud models from a different time. These difference could be used for safety and stability monitoring. Since the point cloud has the ability to remotely (safely), rapidly and accurately extract large quantities of georeferenced and 3D-oriented data, point cloud technology provides numerous applications to the geomechanical field, and the list of uses is continuously growing. Accuracy is the key requirement. Developing a high accurate registration with the background knowledge of these fields is the future research direction.

\section{Open questions and future direction}
Based on the above literature review and application review, the open questions are two folds: (1) high accurate and robust registration by overcoming same-source and cross-source challenges. (2) the fast running speed with the guarantee of high accuracy. In this section, we suggest four future research directions. 
\subsection{Robust and accurate registration}
The point cloud is a record of the 3D environment. However, the real data is much complicated because of the noise and outliers variations. These variations could come from sensors or environment change during the different acquisition time. Firstly, the future direction could be robust to handle the challenging variations of noise and outliers in real-world point clouds. Although many methods are focusing on this area \cite{bustos2017guaranteed, yang2019polynomial,bouaziz2013sparse, huang2017systematic}, both the accuracy and speed are far behind the requirement of real applications. Secondly, high accuracy is another critical research direction. High accuracy is indispensable for many real applications, such as geography survey, high definition map for autonomous driving as discussed in section \ref{applications}. Although the recent deep learning methods can achieve high registration accuracy on the KITTI dataset, e.g. \cite{choy2020deep} obtains 3cm in the KITTI dataset, the robustness and generalization ability to other datasets are still less reported. Thirdly, the generalization ability of learning-based to the real diverse applications is still a remaining research question.

\subsection{Efficiency}
Registration efficiency is another remaining research problem, which also is a future research direction. The recent point clouds usually contain millions of points; the conventional optimization method such as ICP will be extremely slow. However, many current advanced methods are all required ICP to do the refinement to obtain high accuracy. Without the refinement, the accuracy will drop highly. For example, DGR \cite{choy2020deep} obtains 3cm registration accuracy with ICP refinement while the accuracy drops to 22cm without ICP in the KITTI dataset. 

\subsection{Partial overlap}
Partial overlap means only part of the point clouds describe the same 3D environment while the other parts are different. The partial overlap ratio could be very small such as less than 20\%. This overlap ratio will be very challenging since the search for overlap ratio is a combination problem even though our human requires much time to manually align two partial overlapped point clouds to find the common regions. Recent technologies \cite{wang2019prnet, gross2019alignnet} propose keypoint-based solutions to solve partial overlap. They highly rely on the quality of keypoint detection. The future research direction is to design a robust algorithm to solve the low overlapped point cloud registration.

\subsection{Fusion of deep learning and registration mathematical theories}
Many existing experiments \cite{gojcic2019perfect, bes1992method, huang2017systematic} show that directly apply the mathematical theories of registration will cost huge computation time, while directly apply deep learning will not guarantee accuracy. Directly combining deep learning and ICP still require high computation time. Recently, several pieces of literature \cite{aoki2019pointnetlk, wang2019prnet, huang2020feature} are trying to merge the conventional mathematical theories and deep neural network into an end-to-end framework in order to obtain both high accuracy and efficiency. This area is just the beginning, and there needs much research to develop fantastic fusion registration algorithms.

\section{Conclusion}
This paper conducts a comprehensive survey for point cloud registration from same-source and cross-source domains. In this survey, for the first time, we conduct a review of cross-source point cloud registration and evaluate the existing state-of-the-art registration methods on the cross-source dataset. Besides, we summarize the connections between optimization-based and deep learning methods. After that, we summarize the possible applications of point cloud registration. Finally, we propose several future research directions and open questions in the registration field.

{\small
	\bibliographystyle{ieee_fullname}
	\bibliography{egbib}

\begin{thebibliography}{100}\itemsep=-1pt

\bibitem{aguilar20173d}
Fernando~J AGUILAR, Ismael FERN{\'A}NDEZ, Juan~A CASANOVA, Francisco~J RAMOS,
  Manuel~A AGUILAR, Jos{\'e}~L BLANCO, and Jos{\'e}~C MORENO.
\newblock 3d coastal monitoring from very dense uav-based photogrammetric point
  clouds.
\newblock In {\em Advances on Mechanics, Design Engineering and Manufacturing},
  pages 879--887. Springer, 2017.

\bibitem{almohamad1993linear}
HA Almohamad and Salih~O Duffuaa.
\newblock A linear programming approach for the weighted graph matching
  problem.
\newblock {\em IEEE Transactions on pattern analysis and machine intelligence},
  15(5):522--525, 1993.

\bibitem{aoki2019pointnetlk}
Yasuhiro Aoki, Hunter Goforth, Rangaprasad~Arun Srivatsan, and Simon Lucey.
\newblock Pointnetlk: Robust \& efficient point cloud registration using
  pointnet.
\newblock In {\em Proceedings of the IEEE Conference on Computer Vision and
  Pattern Recognition}, pages 7163--7172, 2019.

\bibitem{baker2004lucas}
Simon Baker and Iain Matthews.
\newblock Lucas-kanade 20 years on: A unifying framework.
\newblock {\em International journal of computer vision}, 56(3):221--255, 2004.

\bibitem{Bernard_2018_CVPR}
Florian Bernard, Christian Theobalt, and Michael Moeller.
\newblock Ds*: Tighter lifting-free convex relaxations for quadratic matching
  problems.
\newblock In {\em The IEEE Conference on Computer Vision and Pattern
  Recognition (CVPR)}, June 2018.

\bibitem{bes1992method}
Paul~J Bes, Neil~D McKay, et~al.
\newblock A method for registration of 3-d shapes.
\newblock {\em IEEE Transactions on Pattern Analysis and Machine Intelligence},
  14(2):239--256, 1992.

\bibitem{billings2015iterative}
Seth~D Billings, Emad~M Boctor, and Russell~H Taylor.
\newblock Iterative most-likely point registration (imlp): a robust algorithm
  for computing optimal shape alignment.
\newblock {\em PloS one}, 10(3), 2015.

\bibitem{bishop2006pattern}
Christopher~M Bishop.
\newblock {\em Pattern recognition and machine learning}.
\newblock springer, 2006.

\bibitem{bouaziz2013sparse}
Sofien Bouaziz, Andrea Tagliasacchi, and Mark Pauly.
\newblock Sparse iterative closest point.
\newblock In {\em Computer graphics forum}, volume~32, pages 113--123. Wiley
  Online Library, 2013.

\bibitem{brenner2008coarse}
C Brenner, C Dold, and N Ripperda.
\newblock Coarse orientation of terrestrial laser scans in urban environments.
\newblock {\em ISPRS journal of photogrammetry and remote sensing},
  63(1):4--18, 2008.

\bibitem{bustos2017guaranteed}
{\'A}lvaro~Parra Bustos and Tat-Jun Chin.
\newblock Guaranteed outlier removal for point cloud registration with
  correspondences.
\newblock {\em IEEE transactions on pattern analysis and machine intelligence},
  40(12):2868--2882, 2017.

\bibitem{chen2013point}
Jyun-Yuan Chen, Chao-Hung Lin, Po-Chi Hsu, and Chung-Hao Chen.
\newblock Point cloud encoding for 3d building model retrieval.
\newblock {\em IEEE transactions on multimedia}, 16(2):337--345, 2013.

\bibitem{chen20203d}
Siheng Chen, Baoan Liu, Chen Feng, Carlos Vallespi-Gonzalez, and Carl
  Wellington.
\newblock 3d point cloud processing and learning for autonomous driving.
\newblock {\em arXiv preprint arXiv:2003.00601}, 2020.

\bibitem{chen1992object}
Yang Chen and G{\'e}rard Medioni.
\newblock Object modelling by registration of multiple range images.
\newblock {\em Image and vision computing}, 10(3):145--155, 1992.

\bibitem{cheng2018registration}
Liang Cheng, Song Chen, Xiaoqiang Liu, Hao Xu, Yang Wu, Manchun Li, and Yanming
  Chen.
\newblock Registration of laser scanning point clouds: A review.
\newblock {\em Sensors}, 18(5):1641, 2018.

\bibitem{choy2020deep}
Christopher Choy, Wei Dong, and Vladlen Koltun.
\newblock Deep global registration.
\newblock In {\em Proceedings of the IEEE Conference on Computer Vision and
  Pattern Recognition}, 2020.

\bibitem{choy2019fully}
Christopher Choy, Jaesik Park, and Vladlen Koltun.
\newblock Fully convolutional geometric features.
\newblock In {\em Proceedings of the IEEE International Conference on Computer
  Vision}, pages 8958--8966, 2019.

\bibitem{deng2018ppf}
Haowen Deng, Tolga Birdal, and Slobodan Ilic.
\newblock Ppf-foldnet: Unsupervised learning of rotation invariant 3d local
  descriptors.
\newblock In {\em Proceedings of the European Conference on Computer Vision
  (ECCV)}, pages 602--618, 2018.

\bibitem{deng2018ppfnet}
Haowen Deng, Tolga Birdal, and Slobodan Ilic.
\newblock Ppfnet: Global context aware local features for robust 3d point
  matching.
\newblock In {\em Proceedings of the IEEE Conference on Computer Vision and
  Pattern Recognition}, pages 195--205, 2018.

\bibitem{deng20193d}
Haowen Deng, Tolga Birdal, and Slobodan Ilic.
\newblock 3d local features for direct pairwise registration.
\newblock In {\em Proceedings of the IEEE Conference on Computer Vision and
  Pattern Recognition}, pages 3244--3253, 2019.

\bibitem{drost2010model}
Bertram Drost, Markus Ulrich, Nassir Navab, and Slobodan Ilic.
\newblock Model globally, match locally: Efficient and robust 3d object
  recognition.
\newblock In {\em 2010 IEEE computer society conference on computer vision and
  pattern recognition}, pages 998--1005. Ieee, 2010.

\bibitem{dryden2016statistical}
Ian~L Dryden and Kanti~V Mardia.
\newblock {\em Statistical shape analysis: with applications in R}, volume 995.
\newblock John Wiley \& Sons, 2016.

\bibitem{duchenne2011tensor}
Olivier Duchenne, Francis Bach, In-So Kweon, and Jean Ponce.
\newblock A tensor-based algorithm for high-order graph matching.
\newblock {\em IEEE transactions on pattern analysis and machine intelligence},
  33(12):2383--2395, 2011.

\bibitem{dym2017ds++}
Nadav Dym, Haggai Maron, and Yaron Lipman.
\newblock Ds++: a flexible, scalable and provably tight relaxation for matching
  problems.
\newblock {\em arXiv preprint arXiv:1705.06148}, 2017.

\bibitem{eggert1997estimating}
David~W Eggert, Adele Lorusso, and Robert~B Fisher.
\newblock Estimating 3-d rigid body transformations: a comparison of four major
  algorithms.
\newblock {\em Machine vision and applications}, 9(5-6):272--290, 1997.

\bibitem{eguizabal2019procrustes}
Alma Eguizabal, Peter Schreier, and Juergen Schmidt.
\newblock Procrustes registration of two-dimensional statistical shape models
  without correspondences.
\newblock {\em arXiv preprint arXiv:1911.11431}, 2019.

\bibitem{elbaz20173d}
Gil Elbaz, Tamar Avraham, and Anath Fischer.
\newblock 3d point cloud registration for localization using a deep neural
  network auto-encoder.
\newblock In {\em Proceedings of the IEEE Conference on Computer Vision and
  Pattern Recognition}, pages 4631--4640, 2017.

\bibitem{enqvist2009optimal}
Olof Enqvist, Klas Josephson, and Fredrik Kahl.
\newblock Optimal correspondences from pairwise constraints.
\newblock In {\em 2009 IEEE 12th international conference on computer vision},
  pages 1295--1302. IEEE, 2009.

\bibitem{evangelidis2014generative}
Georgios~D Evangelidis, Dionyssos Kounades-Bastian, Radu Horaud, and
  Emmanouil~Z Psarakis.
\newblock A generative model for the joint registration of multiple point sets.
\newblock In {\em European Conference on Computer Vision}, pages 109--122.
  Springer, 2014.

\bibitem{fan2016convex}
Jingfan Fan, Jian Yang, Danni Ai, Likun Xia, Yitian Zhao, Xing Gao, and
  Yongtian Wang.
\newblock Convex hull indexed gaussian mixture model (ch-gmm) for 3d point set
  registration.
\newblock {\em Pattern Recognition}, 59:126--141, 2016.

\bibitem{Feyetal2020}
M. Fey, J.~E. Lenssen, C. Morris, J. Masci, and N.~M. Kriege.
\newblock Deep graph matching consensus.
\newblock In {\em International Conference on Learning Representations (ICLR)},
  2020.

\bibitem{fitzgibbon2003robust}
Andrew~W Fitzgibbon.
\newblock Robust registration of 2d and 3d point sets.
\newblock {\em Image and vision computing}, 21(13-14):1145--1153, 2003.

\bibitem{forstner2017efficient}
Wolfgang Forstner and Kourosh Khoshelham.
\newblock Efficient and accurate registration of point clouds with plane to
  plane correspondences.
\newblock In {\em Proceedings of the IEEE International Conference on Computer
  Vision Workshops}, pages 2165--2173, 2017.

\bibitem{garey1979computers}
Michael~R Garey and David~S Johnson.
\newblock {\em Computers and intractability}, volume 174.
\newblock freeman San Francisco, 1979.

\bibitem{gojcic2019perfect}
Zan Gojcic, Caifa Zhou, Jan~D Wegner, and Andreas Wieser.
\newblock The perfect match: 3d point cloud matching with smoothed densities.
\newblock In {\em Proceedings of the IEEE Conference on Computer Vision and
  Pattern Recognition}, pages 5545--5554, 2019.

\bibitem{gross2019alignnet}
Johannes Gro{\ss}, Aljo{\v{s}}a O{\v{s}}ep, and Bastian Leibe.
\newblock Alignnet-3d: Fast point cloud registration of partially observed
  objects.
\newblock In {\em 2019 International Conference on 3D Vision (3DV)}, pages
  623--632. IEEE, 2019.

\bibitem{hayes2015use}
Cathi Hayes and Eric Richie.
\newblock When to use laser scanning in building construction, 2015.

\bibitem{hitz2016state}
Gregory Hitz, Fran{\c{c}}ois Pomerlesau, Francis Colas, and Roland Siegwart.
\newblock State estimation for shore monitoring using an autonomous surface
  vessel.
\newblock In {\em Experimental Robotics}, pages 745--760. Springer, 2016.

\bibitem{huang2019fast}
Xiaoshui Huang, Lixin Fan, Qiang Wu, Jian Zhang, and Chun Yuan.
\newblock Fast registration for cross-source point clouds by using weak
  regional affinity and pixel-wise refinement.
\newblock In {\em 2019 IEEE International Conference on Multimedia and Expo
  (ICME)}, pages 1552--1557. IEEE, 2019.

\bibitem{huang2020feature}
Xiaoshui Huang, Guofeng Mei, and Jian Zhang.
\newblock Feature-metric registration: A fast semi-supervised approach for
  robust point cloud registration without correspondences.
\newblock In {\em Proceedings of the IEEE/CVF Conference on Computer Vision and
  Pattern Recognition}, pages 11366--11374, 2020.

\bibitem{huang2017systematic}
Xiaoshui Huang, Jian Zhang, Lixin Fan, Qiang Wu, and Chun Yuan.
\newblock A systematic approach for cross-source point cloud registration by
  preserving macro and micro structures.
\newblock {\em IEEE Transactions on Image Processing}, 26(7):3261--3276, 2017.

\bibitem{huang2016coarse}
Xiaoshui Huang, Jian Zhang, Qiang Wu, Lixin Fan, and Chun Yuan.
\newblock A coarse-to-fine algorithm for registration in 3d street-view
  cross-source point clouds.
\newblock In {\em 2016 International Conference on Digital Image Computing:
  Techniques and Applications (DICTA)}, pages 1--6. IEEE, 2016.

\bibitem{huang2017coarse}
Xiaoshui Huang, Jian Zhang, Qiang Wu, Lixin Fan, and Chun Yuan.
\newblock A coarse-to-fine algorithm for matching and registration in 3d
  cross-source point clouds.
\newblock {\em IEEE Transactions on Circuits and Systems for Video Technology},
  28(10):2965--2977, 2017.

\bibitem{iglesias2020global}
Jos{\'e}~Pedro Iglesias, Carl Olsson, and Fredrik Kahl.
\newblock Global optimality for point set registration using semidefinite
  programming.
\newblock In {\em Proceedings of the IEEE/CVF Conference on Computer Vision and
  Pattern Recognition}, pages 8287--8295, 2020.

\bibitem{ilci2020high}
Veli Ilci and Charles Toth.
\newblock High definition 3d map creation using gnss/imu/lidar sensor
  integration to support autonomous vehicle navigation.
\newblock {\em Sensors}, 20(3):899, 2020.

\bibitem{johnson1999using}
Andrew~E. Johnson and Martial Hebert.
\newblock Using spin images for efficient object recognition in cluttered 3d
  scenes.
\newblock {\em IEEE Transactions on pattern analysis and machine intelligence},
  21(5):433--449, 1999.

\bibitem{kezurer2015tight}
Itay Kezurer, Shahar~Z Kovalsky, Ronen Basri, and Yaron Lipman.
\newblock Tight relaxation of quadratic matching.
\newblock In {\em Computer Graphics Forum}, volume~34, pages 115--128. Wiley
  Online Library, 2015.

\bibitem{khoshelham2010automated}
Kourosh Khoshelham.
\newblock Automated localization of a laser scanner in indoor environments
  using planar objects.
\newblock In {\em 2010 International Conference on Indoor Positioning and
  Indoor Navigation}, pages 1--7. IEEE, 2010.

\bibitem{khoshelham2016closed}
Kourosh Khoshelham.
\newblock Closed-form solutions for estimating a rigid motion from plane
  correspondences extracted from point clouds.
\newblock {\em ISPRS Journal of Photogrammetry and Remote Sensing}, 114:78--91,
  2016.

\bibitem{khoury2017learning}
Marc Khoury, Qian-Yi Zhou, and Vladlen Koltun.
\newblock Learning compact geometric features.
\newblock In {\em Proceedings of the IEEE International Conference on Computer
  Vision}, pages 153--161, 2017.

\bibitem{kim2018automated}
Pileun Kim, Jingdao Chen, and Yong~K Cho.
\newblock Automated point cloud registration using visual and planar features
  for construction environments.
\newblock {\em Journal of Computing in Civil Engineering}, 32(2):04017076,
  2018.

\bibitem{kim2017automatic}
Pileun Kim and Yong~K Cho.
\newblock An automatic robust point cloud registration on construction sites.
\newblock In {\em Computing in Civil Engineering 2017}, pages 411--419. 2017.

\bibitem{kolmogorov2006convergent}
Vladimir Kolmogorov.
\newblock Convergent tree-reweighted message passing for energy minimization.
\newblock {\em IEEE transactions on pattern analysis and machine intelligence},
  28(10):1568--1583, 2006.

\bibitem{le2019sdrsac}
Huu~M Le, Thanh-Toan Do, Tuan Hoang, and Ngai-Man Cheung.
\newblock Sdrsac: Semidefinite-based randomized approach for robust point cloud
  registration without correspondences.
\newblock In {\em Proceedings of the IEEE Conference on Computer Vision and
  Pattern Recognition}, pages 124--133, 2019.

\bibitem{le2017alternating}
D~Khu{\^e} L{\^e}-Huu and Nikos Paragios.
\newblock Alternating direction graph matching.
\newblock In {\em 2017 IEEE Conference on Computer Vision and Pattern
  Recognition (CVPR)}, pages 4914--4922. IEEE, 2017.

\bibitem{lee2019analyzing}
Seung-Joong Lee and Sung-Oong Choi.
\newblock Analyzing the stability of underground mines using 3d point cloud
  data and discontinuum numerical analysis.
\newblock {\em Sustainability}, 11(4):945, 2019.

\bibitem{leordeanu2005spectral}
Marius Leordeanu and Martial Hebert.
\newblock A spectral technique for correspondence problems using pairwise
  constraints.
\newblock In {\em Tenth IEEE International Conference on Computer Vision
  (ICCV'05) Volume 1}, volume~2, pages 1482--1489. IEEE, 2005.

\bibitem{li2021point}
Jiayuan Li, Qingwu Hu, and Mingyao Ai.
\newblock Point cloud registration based on one-point ransac and
  scale-annealing biweight estimation.
\newblock {\em IEEE Transactions on Geoscience and Remote Sensing}, 2021.

\bibitem{li2019iterative}
Jiahao Li, Changhao Zhang, Ziyao Xu, Hangning Zhou, and Chi Zhang.
\newblock Iterative distance-aware similarity matrix convolution with
  mutual-supervised point elimination for efficient point cloud registration.
\newblock In {\em European Conference on Computer Vision}, 2020.

\bibitem{liu2020deformation}
Maohua Liu, Xiubo Sun, Yan Wang, Yue Shao, and Yingchun You.
\newblock Deformation measurement of highway bridge head based on mobile tls
  data.
\newblock {\em IEEE Access}, 8:85605--85615, 2020.

\bibitem{livi2013graph}
Lorenzo Livi and Antonello Rizzi.
\newblock The graph matching problem.
\newblock {\em Pattern Analysis and Applications}, 16(3):253--283, 2013.

\bibitem{loiola2007survey}
Eliane~Maria Loiola, Nair Maria~Maia de Abreu, Paulo~Oswaldo Boaventura-Netto,
  Peter Hahn, and Tania Querido.
\newblock A survey for the quadratic assignment problem.
\newblock {\em European journal of operational research}, 176(2):657--690,
  2007.

\bibitem{low2004linear}
Kok-Lim Low.
\newblock Linear least-squares optimization for point-to-plane icp surface
  registration.
\newblock {\em Chapel Hill, University of North Carolina}, 4(10):1--3, 2004.

\bibitem{lu2019deepicp}
Weixin Lu, Guowei Wan, Yao Zhou, Xiangyu Fu, Pengfei Yuan, and Shiyu Song.
\newblock Deepicp: An end-to-end deep neural network for 3d point cloud
  registration.
\newblock {\em arXiv preprint arXiv:1905.04153}, 2019.

\bibitem{lu2019deepvcp}
Weixin Lu, Guowei Wan, Yao Zhou, Xiangyu Fu, Pengfei Yuan, and Shiyu Song.
\newblock Deepvcp: An end-to-end deep neural network for point cloud
  registration.
\newblock In {\em Proceedings of the IEEE International Conference on Computer
  Vision}, pages 12--21, 2019.

\bibitem{lyons2016applications}
J Lyons-Baral and J Kemeny.
\newblock Applications of point cloud technology in geomechanical
  characterization, analysis and predictive modeling.
\newblock {\em Mining Engineering}, 68(5):18--29, 2016.

\bibitem{maron2016point}
Haggai Maron, Nadav Dym, Itay Kezurer, Shahar Kovalsky, and Yaron Lipman.
\newblock Point registration via efficient convex relaxation.
\newblock {\em ACM Transactions on Graphics (TOG)}, 35(4):1--12, 2016.

\bibitem{mellado2014super}
Nicolas Mellado, Dror Aiger, and Niloy~J Mitra.
\newblock Super 4pcs fast global pointcloud registration via smart indexing.
\newblock In {\em Computer Graphics Forum}, volume~33, pages 205--215. Wiley
  Online Library, 2014.

\bibitem{mellado2015relative}
Nicolas Mellado, Matteo Dellepiane, and Roberto Scopigno.
\newblock Relative scale estimation and 3d registration of multi-modal geometry
  using growing least squares.
\newblock {\em IEEE transactions on visualization and computer graphics},
  22(9):2160--2173, 2015.

\bibitem{monsalve2018preliminary}
J Monsalve, Jon Baggett, Richard Bishop, and Nino Ripepi.
\newblock A preliminary investigation for characterization and modeling of
  structurally controlled underground limestone mines by integrating laser
  scanning with discrete element modeling.
\newblock In {\em North American Tunneling Conference}, 2018.

\bibitem{monsalve2019application}
Juan~J Monsalve, Jon Baggett, Richard Bishop, and Nino Ripepi.
\newblock Application of laser scanning for rock mass characterization and
  discrete fracture network generation in an underground limestone mine.
\newblock {\em International Journal of Mining Science and Technology},
  29(1):131--137, 2019.

\bibitem{muller2007information}
Meinard M{\"u}ller.
\newblock {\em Information retrieval for music and motion}, volume~2.
\newblock Springer, 2007.

\bibitem{myronenko2010point}
Andriy Myronenko and Xubo Song.
\newblock Point set registration: Coherent point drift.
\newblock {\em IEEE transactions on pattern analysis and machine intelligence},
  32(12):2262--2275, 2010.

\bibitem{nagy2018real}
Bal{\'a}zs Nagy and Csaba Benedek.
\newblock Real-time point cloud alignment for vehicle localization in a high
  resolution 3d map.
\newblock In {\em Proceedings of the European Conference on Computer Vision
  (ECCV)}, pages 0--0, 2018.

\bibitem{pais20193dregnet}
G~Dias Pais, Srikumar Ramalingam, Venu~Madhav Govindu, Jacinto~C Nascimento,
  Rama Chellappa, and Pedro Miraldo.
\newblock 3dregnet: A deep neural network for 3d point registration.
\newblock In {\em Proceedings of the IEEE/CVF Conference on Computer Vision and
  Pattern Recognition}, pages 7193--7203, 2020.

\bibitem{pan2019clustermap}
Zhichen Pan, Haoyao Chen, Silin Li, and Yunhui Liu.
\newblock Clustermap building and relocalization in urban environments for
  unmanned vehicles.
\newblock {\em Sensors}, 19(19):4252, 2019.

\bibitem{peng2014street}
Furong Peng, Qiang Wu, Lixin Fan, Jian Zhang, Yu You, Jianfeng Lu, and Jing-Yu
  Yang.
\newblock Street view cross-sourced point cloud matching and registration.
\newblock In {\em 2014 IEEE International Conference on Image Processing
  (ICIP)}, pages 2026--2030. IEEE, 2014.

\bibitem{pomerleau2015review}
Fran{\c{c}}ois Pomerleau, Francis Colas, Roland Siegwart, et~al.
\newblock A review of point cloud registration algorithms for mobile robotics.
\newblock {\em Foundations and Trends{\textregistered} in Robotics},
  4(1):1--104, 2015.

\bibitem{poto2017laser}
Vivien Pot{\'o}, J{\'o}zsef~{\'A}rp{\'a}d Somogyi, Tam{\'a}s Lovas, and
  {\'A}rp{\'a}d Barsi.
\newblock Laser scanned point clouds to support autonomous vehicles.
\newblock {\em Transportation Research Procedia}, 27:531--537, 2017.

\bibitem{qi2017pointnet}
Charles~R Qi, Hao Su, Kaichun Mo, and Leonidas~J Guibas.
\newblock Pointnet: Deep learning on point sets for 3d classification and
  segmentation.
\newblock In {\em Proceedings of the IEEE Conference on Computer Vision and
  Pattern Recognition}, pages 652--660, 2017.

\bibitem{ramalingam2013theory}
Srikumar Ramalingam and Yuichi Taguchi.
\newblock A theory of minimal 3d point to 3d plane registration and its
  generalization.
\newblock {\em International journal of computer vision}, 102(1-3):73--90,
  2013.

\bibitem{rasoulian2012group}
Abtin Rasoulian, Robert Rohling, and Purang Abolmaesumi.
\newblock Group-wise registration of point sets for statistical shape models.
\newblock {\em IEEE transactions on medical imaging}, 31(11):2025--2034, 2012.

\bibitem{riegler2017octnet}
Gernot Riegler, Ali Osman~Ulusoy, and Andreas Geiger.
\newblock Octnet: Learning deep 3d representations at high resolutions.
\newblock In {\em Proceedings of the IEEE Conference on Computer Vision and
  Pattern Recognition}, pages 3577--3586, 2017.

\bibitem{rodriguez2017underground}
G Rodriguez.
\newblock Underground versatile laser scanning solution.
\newblock In {\em Proceedings of the First International Conference on
  Underground Mining Technology}, pages 445--455. Australian Centre for
  Geomechanics, 2017.

\bibitem{rusinkiewicz2001efficient}
Szymon Rusinkiewicz and Marc Levoy.
\newblock Efficient variants of the icp algorithm.
\newblock In {\em Proceedings Third International Conference on 3-D Digital
  Imaging and Modeling}, pages 145--152. IEEE, 2001.

\bibitem{rusu2009fast}
Radu~Bogdan Rusu, Nico Blodow, and Michael Beetz.
\newblock Fast point feature histograms (fpfh) for 3d registration.
\newblock In {\em 2009 IEEE international conference on robotics and
  automation}, pages 3212--3217. IEEE, 2009.

\bibitem{saiti2020application}
Evdokia Saiti and Theoharis Theoharis.
\newblock An application independent review of multimodal 3d registration
  methods.
\newblock {\em Computers \& Graphics}, 91:153--178, 2020.

\bibitem{salti2014shot}
Samuele Salti, Federico Tombari, and Luigi Di~Stefano.
\newblock Shot: Unique signatures of histograms for surface and texture
  description.
\newblock {\em Computer Vision and Image Understanding}, 125:251--264, 2014.

\bibitem{schellewald2005probabilistic}
Christian Schellewald and Christoph Schn{\"o}rr.
\newblock Probabilistic subgraph matching based on convex relaxation.
\newblock In {\em International Workshop on Energy Minimization Methods in
  Computer Vision and Pattern Recognition}, pages 171--186. Springer, 2005.

\bibitem{segal2009generalized}
Aleksandr Segal, Dirk Haehnel, and Sebastian Thrun.
\newblock Generalized-icp.
\newblock In {\em Robotics: science and systems}, volume~2, page 435. Seattle,
  WA, 2009.

\bibitem{tache2009magnebike}
Fabien T{\^a}che, Wolfgang Fischer, Gilles Caprari, Roland Siegwart, Roland
  Moser, and Francesco Mondada.
\newblock Magnebike: A magnetic wheeled robot with high mobility for inspecting
  complex-shaped structures.
\newblock {\em Journal of Field Robotics}, 26(5):453--476, 2009.

\bibitem{tatarchenko2017octree}
Maxim Tatarchenko, Alexey Dosovitskiy, and Thomas Brox.
\newblock Octree generating networks: Efficient convolutional architectures for
  high-resolution 3d outputs.
\newblock In {\em Proceedings of the IEEE International Conference on Computer
  Vision}, pages 2088--2096, 2017.

\bibitem{tombari2010unique}
Federico Tombari, Samuele Salti, and Luigi Di~Stefano.
\newblock Unique shape context for 3d data description.
\newblock In {\em Proceedings of the ACM workshop on 3D object retrieval},
  pages 57--62, 2010.

\bibitem{torr2003solving}
Philip~HS Torr.
\newblock Solving markov random fields using semi definite programming.
\newblock In {\em AISTATS}, pages 1--8, 2003.

\bibitem{turner2016uavs}
Ian~L Turner, Mitchell~D Harley, and Christopher~D Drummond.
\newblock Uavs for coastal surveying.
\newblock {\em Coastal Engineering}, 114:19--24, 2016.

\bibitem{valsesia2020learning}
Diego Valsesia, Giulia Fracastoro, and Enrico Magli.
\newblock Learning localized representations of point clouds with
  graph-convolutional generative adversarial networks.
\newblock {\em IEEE Transactions on Multimedia}, 23:402--414, 2020.

\bibitem{wang2019non}
Lingjing Wang, Jianchun Chen, Xiang Li, and Yi Fang.
\newblock Non-rigid point set registration networks.
\newblock {\em arXiv preprint arXiv:1904.01428}, 2019.

\bibitem{wang2017cnn}
Peng-Shuai Wang, Yang Liu, Yu-Xiao Guo, Chun-Yu Sun, and Xin Tong.
\newblock O-cnn: Octree-based convolutional neural networks for 3d shape
  analysis.
\newblock {\em ACM Transactions on Graphics (TOG)}, 36(4):1--11, 2017.

\bibitem{wang2019deep}
Yue Wang and Justin~M Solomon.
\newblock Deep closest point: Learning representations for point cloud
  registration.
\newblock In {\em Proceedings of the IEEE International Conference on Computer
  Vision}, pages 3523--3532, 2019.

\bibitem{wang2019prnet}
Yue Wang and Justin~M Solomon.
\newblock Prnet: Self-supervised learning for partial-to-partial registration.
\newblock In {\em Advances in Neural Information Processing Systems}, pages
  8814--8826, 2019.

\bibitem{whelan2012kintinuous}
Thomas Whelan, Michael Kaess, Maurice Fallon, Hordur Johannsson, John Leonard,
  and John McDonald.
\newblock Kintinuous: Spatially extended kinectfusion.
\newblock 2012.

\bibitem{wu2011visualsfm}
Changchang Wu et~al.
\newblock Visualsfm: A visual structure from motion system.
\newblock 2011.

\bibitem{wu20153d}
Zhirong Wu, Shuran Song, Aditya Khosla, Fisher Yu, Linguang Zhang, Xiaoou Tang,
  and Jianxiong Xiao.
\newblock 3d shapenets: A deep representation for volumetric shapes.
\newblock In {\em Proceedings of the IEEE conference on computer vision and
  pattern recognition}, pages 1912--1920, 2015.

\bibitem{yang2019polynomial}
Heng Yang and Luca Carlone.
\newblock A polynomial-time solution for robust registration with extreme
  outlier rates.
\newblock {\em arXiv preprint arXiv:1903.08588}, 2019.

\bibitem{yang2020teaser}
Heng Yang, Jingnan Shi, and Luca Carlone.
\newblock Teaser: Fast and certifiable point cloud registration.
\newblock {\em arXiv preprint arXiv:2001.07715}, 2020.

\bibitem{yang2015go}
Jiaolong Yang, Hongdong Li, Dylan Campbell, and Yunde Jia.
\newblock Go-icp: A globally optimal solution to 3d icp point-set registration.
\newblock {\em IEEE transactions on pattern analysis and machine intelligence},
  38(11):2241--2254, 2015.

\bibitem{yang2020learning}
Jiaqi Yang, Chen Zhao, Ke Xian, Angfan Zhu, and Zhiguo Cao.
\newblock Learning to fuse local geometric features for 3d rigid data matching.
\newblock {\em Information Fusion}, 2020.

\bibitem{yang2020color}
Yang Yang, Weile Chen, Muyi Wang, Dexing Zhong, and Shaoyi Du.
\newblock Color point cloud registration based on supervoxel correspondence.
\newblock {\em IEEE Access}, 8:7362--7372, 2020.

\bibitem{Yang_2019_CVPR}
Zhenpei Yang, Jeffrey~Z. Pan, Linjie Luo, Xiaowei Zhou, Kristen Grauman, and
  Qixing Huang.
\newblock Extreme relative pose estimation for rgb-d scans via scene
  completion.
\newblock In {\em The IEEE Conference on Computer Vision and Pattern
  Recognition (CVPR)}, June 2019.

\bibitem{yew20183dfeat}
Zi~Jian Yew and Gim~Hee Lee.
\newblock 3dfeat-net: Weakly supervised local 3d features for point cloud
  registration.
\newblock In {\em European Conference on Computer Vision}, pages 630--646.
  Springer, 2018.

\bibitem{Yew_2020_CVPR}
Zi~Jian Yew and Gim~Hee Lee.
\newblock Rpm-net: Robust point matching using learned features.
\newblock In {\em The IEEE/CVF Conference on Computer Vision and Pattern
  Recognition (CVPR)}, June 2020.

\bibitem{yuan2020deepgmr}
Wentao Yuan, Benjamin Eckart, Kihwan Kim, Varun Jampani, Dieter Fox, and Jan
  Kautz.
\newblock Deepgmr: Learning latent gaussian mixture models for registration.
\newblock In {\em European Conference on Computer Vision}, pages 733--750.
  Springer, 2020.

\bibitem{zass2008probabilistic}
Ron Zass and Amnon Shashua.
\newblock Probabilistic graph and hypergraph matching.
\newblock In {\em 2008 IEEE Conference on Computer Vision and Pattern
  Recognition}, pages 1--8. IEEE, 2008.

\bibitem{3dmatch}
Andy Zeng, Shuran Song, Matthias Nie{\ss}ner, Matthew Fisher, Jianxiong Xiao,
  and Thomas Funkhouser.
\newblock 3dmatch: Learning local geometric descriptors from rgb-d
  reconstructions.
\newblock In {\em Proceedings of the IEEE Conference on Computer Vision and
  Pattern Recognition}, pages 1802--1811, 2017.

\bibitem{zhang2012robust}
Dong Zhang, Teng Huang, Guihua Li, and Minwei Jiang.
\newblock Robust algorithm for registration of building point clouds using
  planar patches.
\newblock {\em Journal of Surveying Engineering}, 138(1):31--36, 2012.

\bibitem{zhang2020deep}
Zhiyuan Zhang, Yuchao Dai, and Jiadai Sun.
\newblock Deep learning based point cloud registration: an overview.
\newblock {\em Virtual Reality \& Intelligent Hardware}, 2(3):222--246, 2020.

\bibitem{zhou2015factorized}
Feng Zhou and Fernando De~la Torre.
\newblock Factorized graph matching.
\newblock {\em IEEE transactions on pattern analysis and machine intelligence},
  38(9):1774--1789, 2015.

\bibitem{zhou2020siamesepointnet}
J Zhou, MJ Wang, WD Mao, ML Gong, and XP Liu.
\newblock Siamesepointnet: A siamese point network architecture for learning 3d
  shape descriptor.
\newblock In {\em Computer Graphics Forum}, volume~39, pages 309--321. Wiley
  Online Library, 2020.

\bibitem{zhou2016fast}
Qian-Yi Zhou, Jaesik Park, and Vladlen Koltun.
\newblock Fast global registration.
\newblock In {\em European Conference on Computer Vision}, pages 766--782.
  Springer, 2016.

\bibitem{zhou2018open3d}
Qian-Yi Zhou, Jaesik Park, and Vladlen Koltun.
\newblock Open3d: A modern library for 3d data processing.
\newblock {\em arXiv preprint arXiv:1801.09847}, 2018.

\bibitem{zhu2019elastic}
Hu Zhu, Chunfeng Cui, Lizhen Deng, Ray~CC Cheung, and Hong Yan.
\newblock Elastic net constraint-based tensor model for high-order graph
  matching.
\newblock {\em IEEE Transactions on Cybernetics}, 2019.

\bibitem{zhu2019review}
Hao Zhu, Bin Guo, Ke Zou, Yongfu Li, Ka-Veng Yuen, Lyudmila Mihaylova, and
  Henry Leung.
\newblock A review of point set registration: From pairwise registration to
  groupwise registration.
\newblock {\em Sensors}, 19(5):1191, 2019.

\bibitem{zhu2018lidar}
Qingyuan Zhu, Jinjin Wu, Huosheng Hu, Chunsheng Xiao, and Wei Chen.
\newblock Lidar point cloud registration for sensing and reconstruction of
  unstructured terrain.
\newblock {\em Applied Sciences}, 8(11):2318, 2018.

\end{thebibliography}
}

%\newpage
%\section{Appendix 1: Semi-definite formulation}

\end{document}